\documentclass[12pt,a4paper]{article}
\usepackage[english]{babel}

\usepackage[a4paper, top=3cm,bottom=3cm,left=3cm,right=3cm,marginparwidth=1.75cm]{geometry}
\usepackage{amsmath,amsfonts,amsthm}
\usepackage{natbib}
\usepackage{graphicx}
\usepackage{mathtools}
\usepackage[colorlinks=true, allcolors=blue]{hyperref}
\usepackage{xurl}
\usepackage{todonotes}
\usepackage{setspace}

\title{Actuarial Learning for Pension Fund Mortality Forecasting}

\author{Eduardo Fraga L. de Melo$^{1,2}$, Helton Graziadei$^1$ e Rodrigo Targino$^1$}

\date{ }

\begin{document}

\maketitle
\ \ \ \ \ \ \ \ \ \ \ \ \ \ \ \ \ \ {\footnotesize 1 EMAp - Escola de Matemática Aplicada / FGV}

\ \ \ \ \ \ \ \ \ \ \ \ \ \ \ \ \ \ \ \ {\footnotesize 2 UERJ - Univ. do Estado do Rio de Janeiro}

\begin{abstract}

\noindent For the assessment of the financial soundness of a pension fund, it is necessary to take into account mortality forecasting so that longevity risk is consistently incorporated into future cash flows. In this article, we employ machine learning models applied to actuarial science ({\it actuarial learning}) to make mortality predictions for a relevant sample of pension funds' participants. Actuarial learning represents an emerging field that involves the application of machine learning (ML) and artificial intelligence (AI) techniques in actuarial science. This encompasses the use of algorithms and computational models to analyze large sets of actuarial data, such as regression trees, random forest, boosting, XGBoost, CatBoost, and neural networks (eg. FNN, LSTM, and MHA). Our results indicate that some ML/AI algorithms present competitive out-of-sample performance when compared to the classical Lee-Carter model. This may indicate interesting alternatives for consistent liability evaluation and effective pension fund risk management.

\ \ 


\end{abstract}

\newpage

\section{Introduction}

\onehalfspace

Consistent and realistic evaluation of pension fund liabilities demands updated mortality rates and also the forecasting of these rates for future cash flow discounting. These requisites align with contemporary solvency assessment principles and financial reporting standards (refer to \cite{sandstrom2016} and \cite{iasb2017}). Even though pension funds represent the interest of hundreds of millions across the globe, the majority of studies projecting mortality rates rely on national population data (e.g., \cite{cairns2009}, \cite{dowd2011}, \cite{lee1992}, \cite{li2013}, \cite{renshaw2006}, \cite{li2005}).

While using national mortality rates might not pose a significant issue in regions where population mortality is relatively homogeneous, it may introduce basis risk when disparities exist between the national population and a target group. Developed countries often exhibit more homogeneous mortality patterns, but in nations marked by significant social disparities, such as varying income levels and access to healthcare, mortality rates can diverge notably between the general population and a specific group, like pension fund participants.

Brazil, a country characterized by social inequalities (\cite{ocde2018}), illustrates this point. National mortality rates, reflecting the entire population, exceed those of selected groups like pension fund participants. According to regulatory bodies, fewer than 8\%\footnote{\url{https://www.investidorinstitucional.com.br/sessoes/investidores/seguradoras/40265-previdencia-aberta-atinge-5-3-da-populacao-brasileira-em-maio.html} - {\it in portuguese}} of Brazilians possess a pension plan (open or closed), primarily concentrated among the higher income segment. Our analysis indicates that, on average, mortality rates among pensioners are 68\% of those observed in the broader Brazilian population for ages 30 and above.

In this study, interest lies in ages relevant to pension plans, particularly those above 30 years, since, given the nature of our pension fund sample, information for ages under 30 is insufficient. Most participants and retirees exceed this age. In this article, we use machine learning models (most of them implemented in $\mathbf{R}$) within the context of actuarial science, known as actuarial learning, to predict mortality trends in individuals above 30 years old, using data from some pension funds in the Brazilian industry. CatBoost was implemented in $\mathbf{Python}$. The neural network models (feed-forward - FNN, long-short term memory - LSTM, and multi-head attention - MHA) were implemented using $\mathbf{Keras}$ package in $\mathbf{R}$. Our results indicate that the use of actuarial learning models generates consistent results for mortality projection, especially for short projection horizons.

\subsection{Actuarial Learning}

Since its origins, artificial intelligence has been driven by the desire to understand and reveal complex relationships in data, with the aim of developing models capable not only of making accurate predictions but also of extracting knowledge in an understandable way. In this quest, the field of machine learning has diversified considerably, resulting in a wide range of research exploring different aspects and methodologies.

Within the spectrum of machine learning methods, decision tree-based techniques stand out for their effectiveness and usefulness, offering both reliable and interpretable results for a wide variety of datasets. The development of decision trees dates back to \cite{morgan1963}, who introduced the concept through the Automatic Interaction Detector (AID) method, aimed at handling non-additive effects. This initial milestone was followed by significant evolutions and the creation of dedicated computer programs for data analysis, notable contributions made by researchers such as \cite{messenger1972}.

However, the methodological evolution of decision trees was significantly driven by pioneering contributions from \cite{breiman1984}, \cite{friedman1977}, \cite{friedman2001}, and \cite{quinlan1979, quinlan1986}. These researchers substantially enriched the field of machine learning by developing pioneering algorithms for decision trees. The frequent choice of decision trees and related techniques stems from a number of advantageous attributes that position them as highly efficient and accessible analytical tools:

\begin{itemize}
\item Non-parametric nature: they are able to model complex relationships without the need for initial assumptions about the distribution of the data;
\item Flexibility with data types: ability to process heterogeneous data, whether ordinal, categorical, or a combination of both;
\item Intrinsic variable selection: efficiency in identifying and using only the most relevant variables, increasing the model's robustness against irrelevant data;
\item Robustness against outliers and errors: tolerance to anomalies in the data, which contributes to building reasonably stable models;
\item High interpretability: ease of understanding results by users with little or no background in statistics or actuarial science, democratizing access to complex data analysis.
\end{itemize}

It is relevant to emphasize that decision trees form the foundation of a variety of modern algorithms, such as random forests, boosting (\cite{freund1995, freund1997}; \cite{friedman2001}), and XGBoost (\cite{chen2016xgboost}), where they are used as building blocks for more complex models. For a comprehensive overview of various machine learning techniques, one recommends \cite{james2023}. This article also explores the use of neural networks, an artificial intelligence methodology that enables computers to interpret data. Neural networks, a field of machine learning (ML) known as deep learning, are organized into layered structures composed of interconnected nodes or neurons.

In this sense, actuarial learning represents an innovative field that integrates machine learning and artificial intelligence techniques into actuarial science, applying advanced algorithms and computational models for the analysis of vast volumes of actuarial data, including information on insurance, pensions, and financial risks. These methodologies offer broad applications in the actuarial domain, such as:

\begin{itemize}
\item Insurance pricing: use of machine learning algorithms to establish insurance premiums, taking into account a more accurate assessment of customer-associated risks - \cite{Noll2018, Maynard2019}.

\item Risk and claims analysis: use of algorithms to predict risks and identify trends that may signal fraud or claims patterns - \cite{Faheem2022}.

\item Investment and provisioning management: application of predictive models to enhance investment strategies and provision management - \cite{Novykov2023}.

\item Customization of plans: development of personalized insurance plans, based on detailed analyses of each client's specific needs - \cite{Salazar2019}.
\end{itemize}

Thus, the adoption of ML techniques is revolutionizing the way actuaries conduct their analyses, offering deeper insights into risks and behavioral patterns, as well as significantly enhancing strategic decision-making in the insurance, health, and pension sectors. Some notable references in this emerging field include \cite{wuthrichmerz2023} and \cite{denuit2019}.

Within this context, ML emerges as a powerful alternative to Generalized Linear Models (GLMs), for example, accelerating modeling and prediction by identifying complex and often nonlinear data structures, without the need for prior assumptions about the relationship between covariates and the response variable, or about the underlying probability distributions. This approach enables the generation of robust predictive models capable of capturing the complexity of actuarial data, paving the way for unprecedented innovations and efficiencies in the field.

\cite{richmanWutrich2021} show that a deep neural network architecture may outperform the Lee-Carter model considering all countries in the Human Mortality Database (HMD) for mortality rates since 1950. The architecture consists of a covariate (feature) layer, five intermediate layers, and an output layer. The feature layer takes as inputs the year, age, country, and gender of each mortality rate.

\cite{Makhonza2024} considers a modified version of this architecture, where the model is adjusted on unscaled logarithms and the activation function of the output layer is the linear activation function. The approach adopted by \cite{richmanWutrich2021} introduces knowledge of the complete dataset to the model during the training phase by scaling the entire dataset using the minimum and maximum values of the entire dataset.

In this article, we investigate the application of various ML algorithms to predict one-year ahead mortality rates, using covariates in a data-driven methodology. We evaluate and compare the out-of-sample performance of various models, including the Lee-Carter model and ML techniques such as regression trees, random forests, Boosting, XGBoost, and CatBoost, as well as Feedforward Neural Network (FNN), Long-Short Term Memory (LSTM), and Multi-Head Attention (MHA) architectures, to identify those that exhibit the best performance.

\ \ 

\section{Models}

In this section, we briefly describe the ML algorithms used in this study.

\ \

\noindent \textbf{Regression Tree - RT}

\vspace{+0.1in}

Decision trees are commonly used machine learning methods for both classification and regression. They offer popular methods for measuring variable importance and binary separation. For this method, we refer to \cite{breiman1984} and \cite{denuit2019}. Decision trees selects variables to partition the covariate space, and the importance of the variables is measured by analyzing the contribution of each component to the total drop in the objective function (the mean-squared error). A notable advantage of these methods is the ability to perform binary splits additively, simplifying the interpreation of the final model. 

Regression trees form the basis to different regression algorithms, such as random forests and boosting, which are described below.

\ \

\noindent \textbf{Random Forest - RF}

\vspace{+0.1in}

Random Forests, an algorithm proposed by \cite{breiman2001}, offer a strategy for handling interaction and nonlinear effects. The basis of this method lies in developing regression trees through the CART algorithm \citep{breiman1984} of classification and regression trees. This algorithm systematically partitions the predictor variable space through a recursive process, seeking to identify optimal divisions in the training data, minimizing a quadratic loss function.

The process begins by repeatedly sampling with replacement from the training dataset, generating what is known as a bootstrap sample \citep{efron1994}. This bootstrap sample serves as the basis for constructing individual tall trees using the CART algorithm. By repeating this procedure, a collection of independent regression trees is obtained.

To further increase predictive power, these individual trees are aggregated using a technique called bagging \citep{breiman1996}. In bagging, multiple regression functions are trained from bootstrap samples, and their predictions are calculated to create a composite regression function. To optimize this aggregation process, a refinement is introduced. At each decision point within each regression tree in the ensemble, a random subset of predictors is selected without replacement. This enhancement gives rise to what is commonly referred to as a random forest \citep{breiman2001}.

An example of the application of regression trees, random forest, and boosting in actuarial science can be found in \cite{levantesi2019}.

\ \

\noindent \textbf{Boosting}

\vspace{+0.1in}

Similar to the random forest, boosting also consists of aggregating different estimators of the regression function. However, this combination is done differently. There are some variations and implementations of boosting, but the estimator is built incrementally. Initially, it is assigned the value of 0. This estimator has a high bias but low variance (zero). At each step, the value of the estimator is updated to decrease the bias and increase the variance of the new function. This is done by adding to the estimator a function that predicts the residuals.

One way to do this is by using a regression tree and it is important that this tree has a shallow depth to avoid overfitting. Additionally, instead of simply adding this function in full, it is added multiplied by $\lambda$ (called the learning rate): a factor between $0$ and $1$ aimed at preventing overfitting. Another different implementation of boosting mentioned in this article and famous for good performance is XGBoost \citep{chen2016xgboost}.

\ \

\noindent \textbf{XGBoost}

\vspace{+0.1in}

XGBoost (also known as eXtreme Gradient Boosting) is a scalable machine-learning scheme for boosting trees. The predictive capabilities of this algorithm have been widely recognized in a large number of ML challenges and competitions. For this method, we refer to \cite{chen2016xgboost}. XGBoost is designed to be fast, scalable, and highly effective for various machine learning tasks, particularly in predictive modeling and classification problems. The algorithm builds an ensemble of weak prediction models, typically decision trees, in a sequential manner. Each new tree corrects the errors made by the previous ones, aiming to minimize a specified loss function through gradient descent optimization.

\ \

\noindent \textbf{CatBoost}

\vspace{+0.1in}

Gradient boosting is a powerful machine learning technique that achieves great results in a variety of practical tasks. For several years, it has remained the leading method for learning problems with heterogeneous features, noisy data, and complex dependencies. CatBoost (see \cite{catboost2018}) successfully handles categorical features and takes advantage of dealing with them during training, as opposed to preprocessing time. Another advantage of this algorithm is that it uses a new scheme to calculate branch values when selecting the tree structure, which helps reduce overfitting. As a result, the new algorithm may outperform existing state-of-the-art implementations of gradient-boosted decision trees on a diverse set of popular tasks. The algorithm is called CatBoost (for "Categorical Boosting") and has been released as open source.

The CatBoost algorithm uses an efficient strategy that reduces overfitting and allows using the entire dataset for training. In this algorithm, a random permutation of the dataset is performed, and for each example, the average response value is calculated for the example with the same category value placed before that provided in the permutation.

\ \

\noindent \textbf{Feedforward Neural Network (FNN), Long-Short term memory (LSTM) and Multi-Head Attention (MHA)}

\vspace{+0.1in}

In our study, we applied deep learning using FNN, LSTM, and MHA architectures. FNN, consisting of interconnected neurons arranged in successive layers, was trained and validated on the data. Each neuron processes weighted numeric information through an activation function, with the output serving as input for other neurons. FNNs are used for both classification and regression tasks.

The intermediate layers of neurons between the input information and the result are called hidden layers. In this configuration, there is no circular relationship between neurons. Neural networks with circular connections are called recurrent (RNN). FNNs with only one hidden layer are called shallow networks, while FNNs with more than two hidden layers are called deep networks. In actuarial science and mortality forecasting, neural networks have been used in \cite{hainaut2018}, \cite{nigri2019}, and \cite{richman2019}.

RNN is a class of neural networks that can be used to model sequence data (see \cite{denuit2019} and its references). The connections between cells in an RNN form a directed graph over a time sequence, and RNNs use the internal state (memory) of the cells to capture dynamics and temporal dependencies. Long-Short Term Memory (LSTM) \citep{hochreiter1997} is a special type of RNN capable of learning long-term dependencies. An LSTM cell includes two internal states: a cell state that is a vector designed to hold long-term memory and a hidden state that is the output vector of the cell representing the current working memory.

At each time interval, an LSTM cell (Long Short-Term Memory) processes time series data and receives two states from the previous LSTM cell as inputs. Subsequently, it updates its internal states through an input gate, an output gate, and a forget gate. This cell is capable of memorizing values for extensive time intervals, and the mentioned three gates regulate the flow of incoming and outgoing information from the cell, allowing efficient modeling of long-duration dependencies in sequential data. Details of the algorithm can be found in \cite{hochreiter1997}. This feature may have significant relevance and applicability in the field of actuarial science.  This enables the analyst to identify complex patterns in time series of financial and actuarial data, enabling actuaries and risk analysts to make more accurate and informed estimates. Examples of papers using LSTM in actuarial science can be found in \cite{richman2021, nigri2019}. 

More recently, other RNN architectures have been proposed in the literature, in particular by the Natural Language Processing (NLP) community. One of the major breakthroughs was the MHA algorithm from \cite{vaswani2017} which forms one of the building blocks of the Transformer algorithm. This algorithm revolutionized the field due to its ability to handle sequential data, such as sentences or texts, very effectively.  The Transformer stands out for some key features:

\begin{itemize}

\item Self-Attention: this is a fundamental part of the Transformer. It allows the model to analyze a sequence of data in relation to all others, assigning weights to each data according to its importance to the sequence under analysis. This helps the model better capture relationships between the data.

\item Encoders and Decoders: The transformer consists of an encoder and a decoder. The encoder processes the input, and the decoder generates the desired output.

\item Multiple layers of activation: the model consists of several repeated layers, allowing for deeper and more complex analysis of data sequences. Each layer adds more information and refinement to the processing.

\item Activation masks: to ensure the model does not ``see'' information that is not yet available to it during output generation (in the decoder), masks are used to hide parts of the input that should not be used in the current prediction.

\end{itemize}

This model, due to its ability to handle sequences more efficiently and effectively than other models, has become the basis for many state-of-the-art NLP models, including the renowned GPT (Generative Pre-trained Transformer) and BERT (Bidirectional Encoder Representations from Transformers). Additional details on these models are available in \cite{radford2018, devlin2018bert}. These models have revolutionized the NLP field with their capabilities to generate and interpret human language more closely to human understanding. GPT, with its focus on text generation, and BERT, with its ability to understand the bidirectional context of words in a sentence, offer powerful foundations for various applications. In the actuarial context, these models have been utilized in various studies, including in \cite{wang2024, troxler2024}. 

In the actuarial literature there are several approaches for mortality forecasting which use covariates associated with previous ages or years, the age-period models (see \cite{lee1992} and the CBD model - \cite{cairns2006}). Cohort effects are also used by some models (for example \cite{renshawHaberman2006}). In this sense, LSTM and MHA neural networks seem promising since these algorithms use past or proxy information for prediction.

\subsection{Data}

\begin{figure}[!]
\caption{Top row: time series of (i) exposure and (ii) number of deaths per year for the pension fund. Data from 2012 to 2021, both genders and all ages. Second row: Pyramids of (i) exposure and (ii) number of deaths by age and gender in 2021. Bottom row: Raw log of age-specific mortality rates for males and females for years 2012-2021.}
\vspace{+0.1in}

\includegraphics[width=7.35cm]{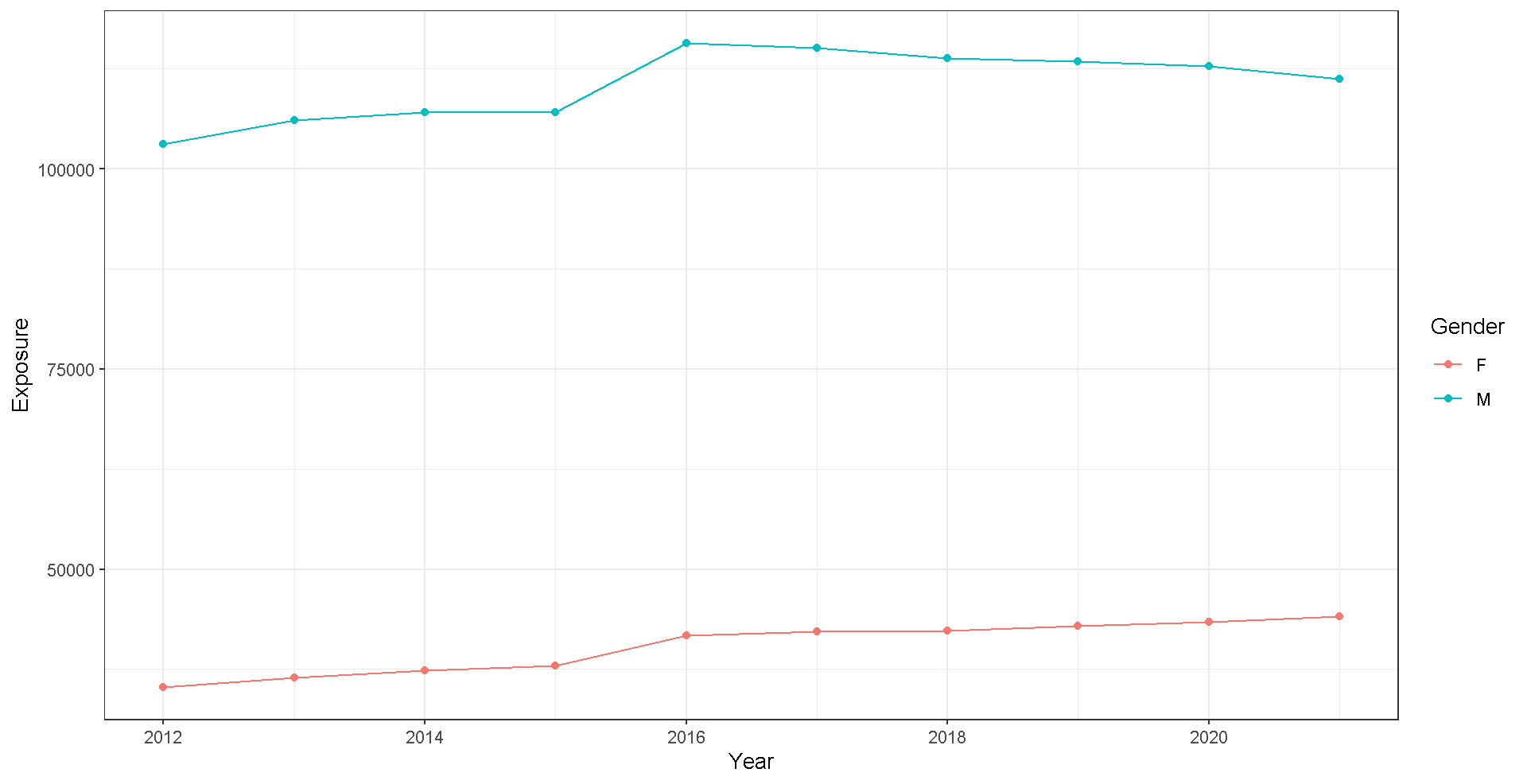}
\includegraphics[width=7.35cm]{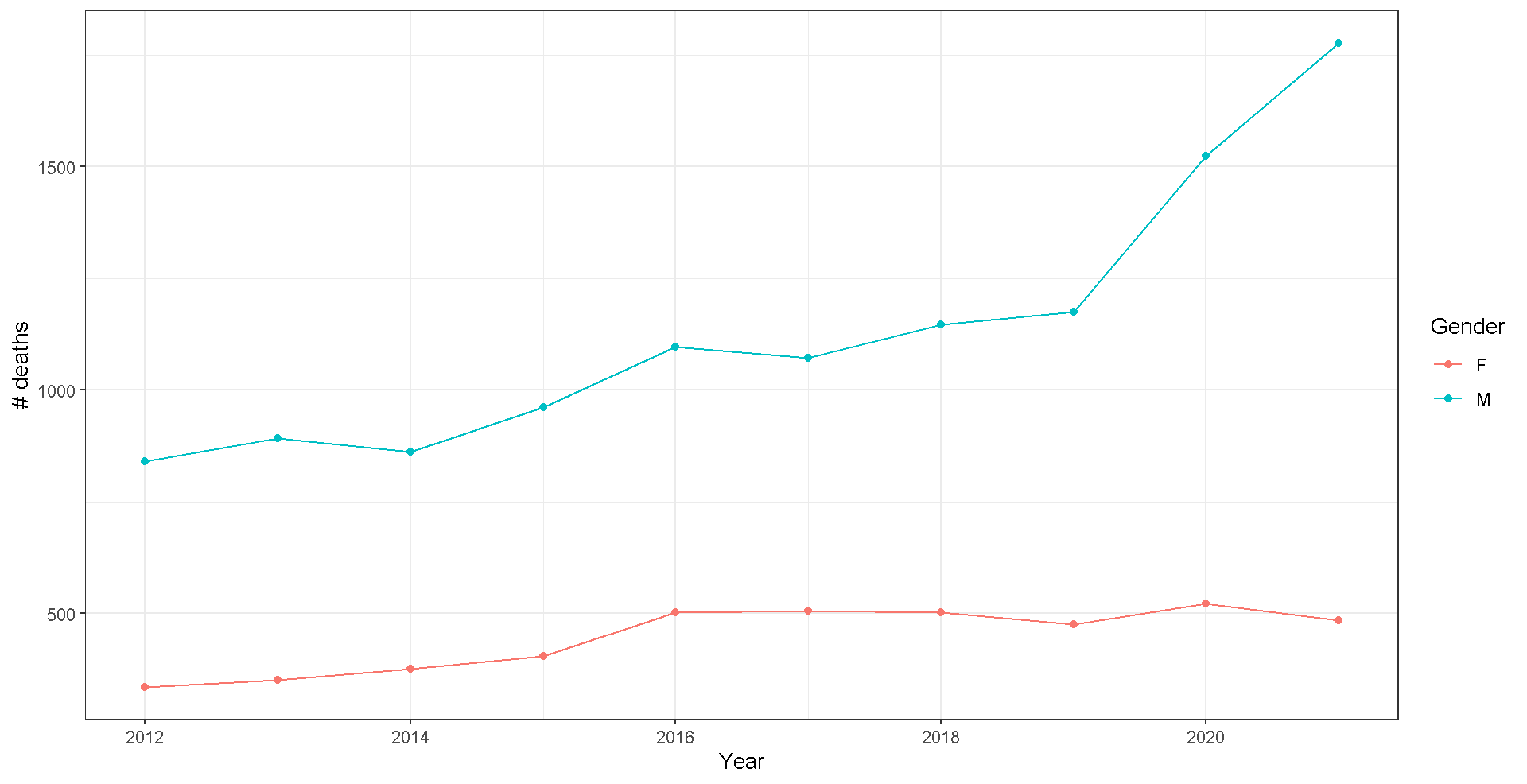}
\includegraphics[width=7.35cm]{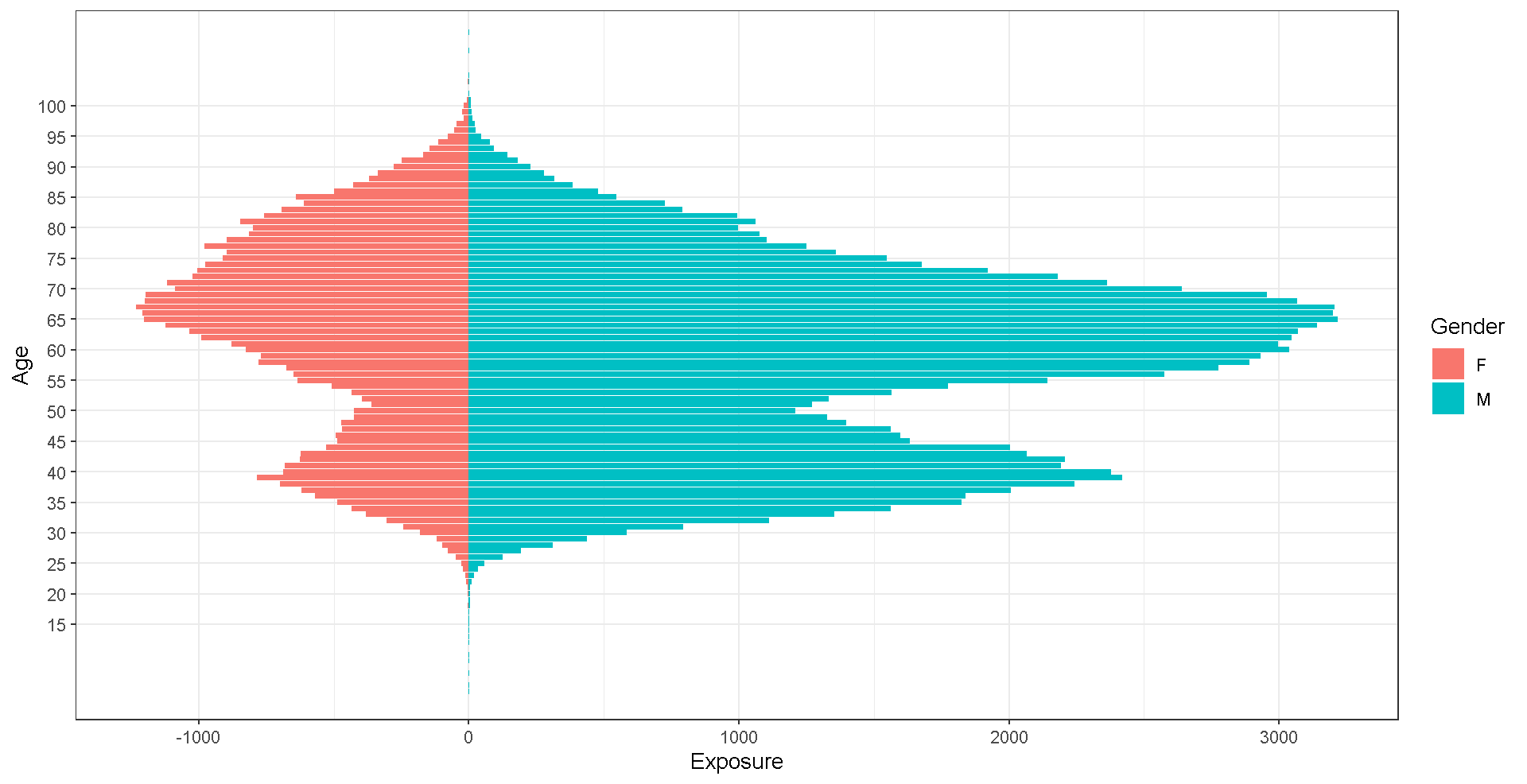}
\includegraphics[width=7.35cm]{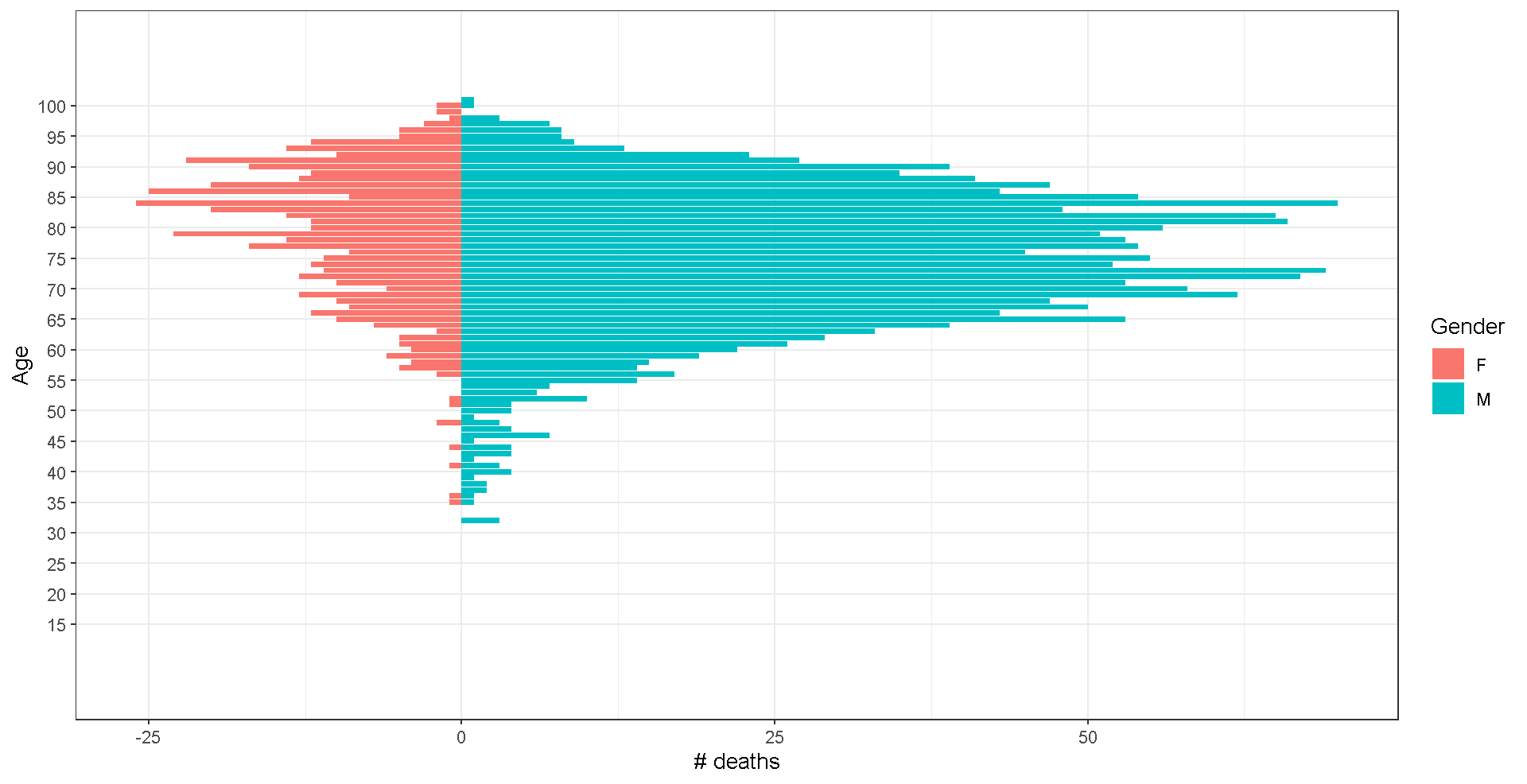}
\includegraphics[width=7.35cm]{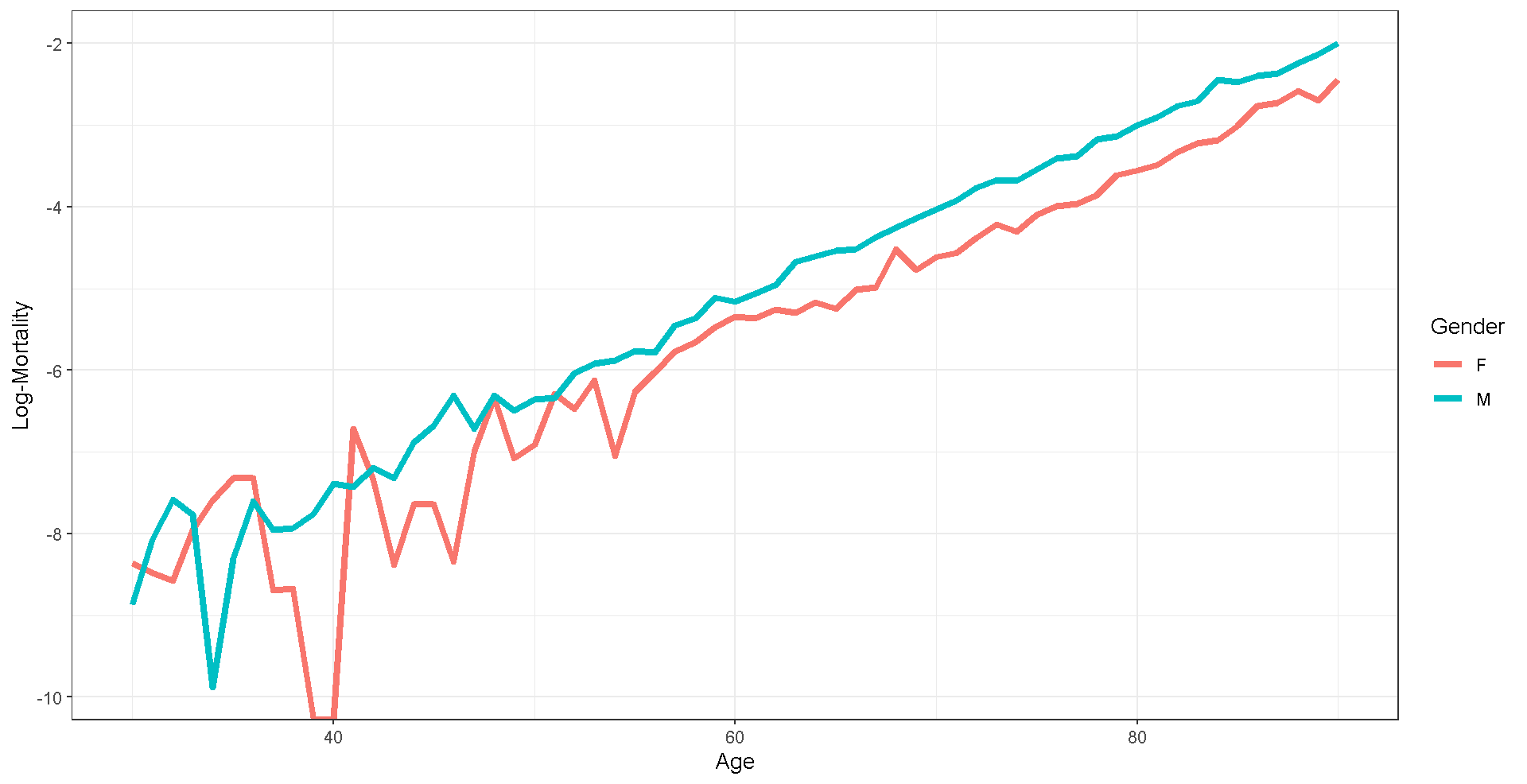}

\label{fig:funds_exp}
\end{figure}

The pension fund data includes exposure and the number of deaths from 2012 to 2021 (10 years), for each gender. Figure \ref{fig:funds_exp} shows the annual time series of total exposure and number of deaths by gender, and the age distribution of these quantities by gender for the year 2021. As noted, exposure and the number of deaths are significantly higher for males.

As one may observe in Figure \ref{fig:funds_exp}, there is an increase in the number of deaths in 2020 and 2021. Possibly, this fact is due to the COVID-19 pandemic. Regarding the age distribution of the exposed population, it reflects a significant number of individuals in the retirement ages, where there is also the addition of pensioners to the sample, who are spouses receiving income due to the death of the principal insured. Last row of Figure \ref{fig:funds_exp} shows the logarithm of mortality rates by gender and year. The plot with curves by gender clearly denotes the difference between male and female mortality - the latter having lower rates.

\subsection{Mortality Forecasting}

\subsubsection{Response Variable and covariates}

In this section, we detail the structure of the response variable and the covariates used in formulating our models. Consider that:

\begin{itemize} 
\item $i$ represents gender, taking values $M$ for male or $F$ for female;
\item $x$ denotes the age, where $x=30,\ldots,95$;
\item $t$ denotes the year, i.e., $t=2012,\ldots,2021$;
\item $D^i_{x,t}$ is the number of deaths at age $x$, calendar year $t$ for gender $i$ for the sample of pension funds;
\item $E^i_{x,t}$ is the exposure at age $x$, calendar year $t$ for gender $i$ for the same funds. 
\end{itemize}

The variable we aim to model, i.e., our response variable, is the mortality rate, defined as:

$$
m^i_{x,t} = \frac{D^i_{x,t}}{E^i_{x,t}}
$$

This rate reflects the proportion of deaths in relation to the exposure for each specified demographic group. To predict these mortality rates using machine learning methods, we selected age ($x$), year ($t$), gender ($i$), and the mortality rate from the previous year ($m^i_{x,t-1}$) as covariates. This approach allows for capturing temporal trends and demographic variations in mortality rates and thus forecasting them.

\subsubsection{Lee-Carter Model}

Machine learning methods do not assume a specific distribution for mortality rates. In addition to machine learning methods, we also apply the model introduced by \cite{lee1992}, as it is a benchmark in mortality forecasting literature, widely used in different versions in academia and practical applications. In this model, the number of deaths for gender $i$ at age $x$ in year $t$ follows:

\begin{equation*}
D^i_{x,t} \sim Poi(e^{\mu^i_{x,t}} E^i_{x,t})
\end{equation*}

The logarithm of the mortality rate ($\mu^i_{x,t}$) has a parametric structure that depends on age and time:

\begin{equation*}
    \mu^i_{x,t} = a^i_x + b^i_x \kappa^i_t
\end{equation*}

\begin{equation*}
    \kappa^i_t = \kappa^i_{t-1} + d^i + \epsilon^i_t
\end{equation*}

For the model to be identifiable, the following conditions must be met: $\sum_x b^i_{x} = 1$ and $\sum_t \kappa^i_t = 0$. In this paper, we performed the Lee-Carter model in a Bayesian approach with the use of the \textbf{StanMoMo} package in \textbf{R} - \cite{stanmomo}.

\subsubsection{Architectures of FNN, LSTM and MHA Networks}

In this subsection, we present the architectures of the networks used in this article. Additionally, we present here the two different ways in which LSTM and MHA networks were used in order to learn about mortality rates by gender, age, and year.

\begin{figure}[!]
\caption{Architectures of the networks (i) FNN and (ii) FNN and LSTM/MHA jointly used for learning mortality rates.}
\vspace{+0.1in}

\centering

\includegraphics[width=10cm]{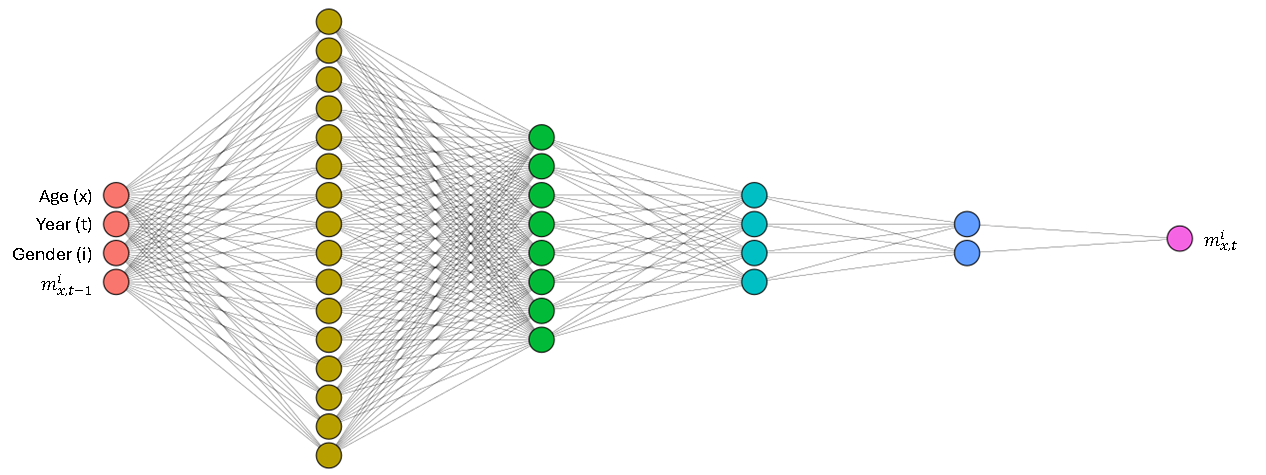}
\vspace{+0.1in}

(i) FNN architecture

\vspace{+0.25in}

\includegraphics[width=10cm]{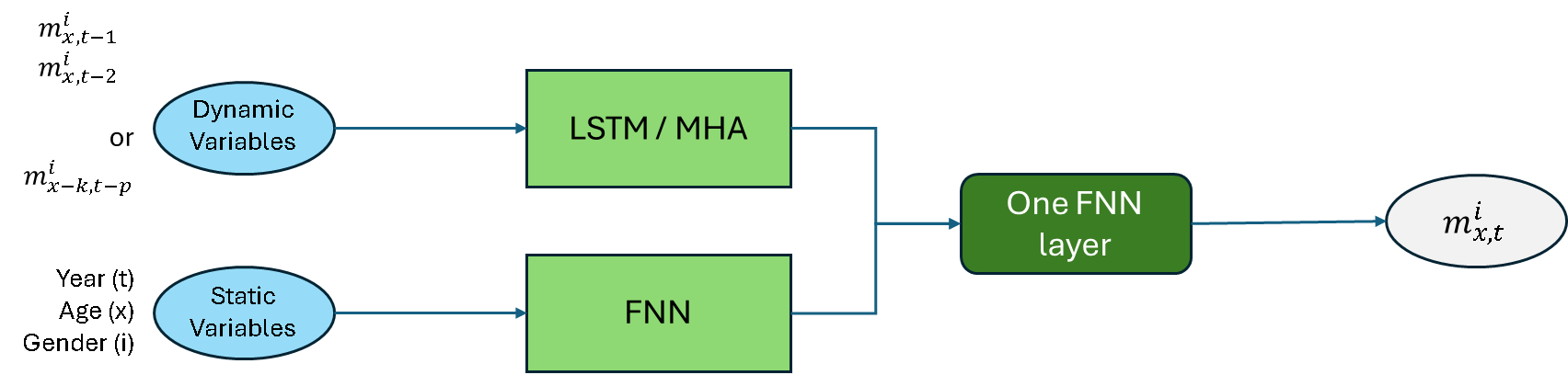}
\vspace{+0.1in}

(ii) FNN architecture combined with LSTM or MHA

\label{fig:arquit}
\end{figure}

The FNN network used in the article has its architecture presented in the plot (i) of Figure \ref{fig:arquit}. The LSTM and MHA networks, on the other hand, were applied with input data organized in two different ways.

The first approach used for LSTM and MHA networks considered the mortality rate data for gender $i$, at age $x$, in year $t$, as dependent on the respective mortality rate of the same age $x$ and gender $i$, but from previous years $t-1$ and $t-2$. The models were called LSTM-1 and MHA-1. The second approach used for LSTM and MHA networks considered the mortality rate data at age $x$ as a long time series, considering jointly age and year. The models were called LSTM-2 and MHA-2.

In order to predict the response variable, composed of the mortality rate $m_{x,t}^i$ for a given age $x$, year $t$, and gender $i$, we define the vector of predictor variables, denoted by $W_{x,t}^i$. This vector is composed of age $x$, year $t$, gender $i$ (where $i = \{0, 1\}$, indicating male if $i = 1$), in addition to a set of mortality rates from previous years and/or ages that are directly considered in the LSTM and MHA architectures. Specifically, to capture the temporal dynamics in mortality rates, in LSTM-1 and MHA-1 models, the mortality rates from two immediately preceding years ($t-1$ and $t-2$) are included for all ages $x$, for both male and female genders. Thus, the vector of $W_{x,t}^i$ is defined as:

$$W_{x,t}^i = [x, i, t, m_{x,t-1}^{i}, m_{x,t-2}^{i}]$$

\begin{figure}[!]
\caption{Illustration of how the mortality rate series ($m^i_{x,t}$) is handled in LSTM-2 and MHA-2 models. The black line denotes the observed mortality rates arranged by Year and Age, between 2012 and 2018, from 30 to 95 years, for Males. The blue dashed line denotes the observed rates for 2019, from 30 to 95 years, for Males.}
\vspace{+0.1in}
\centering

\includegraphics[width=12cm]{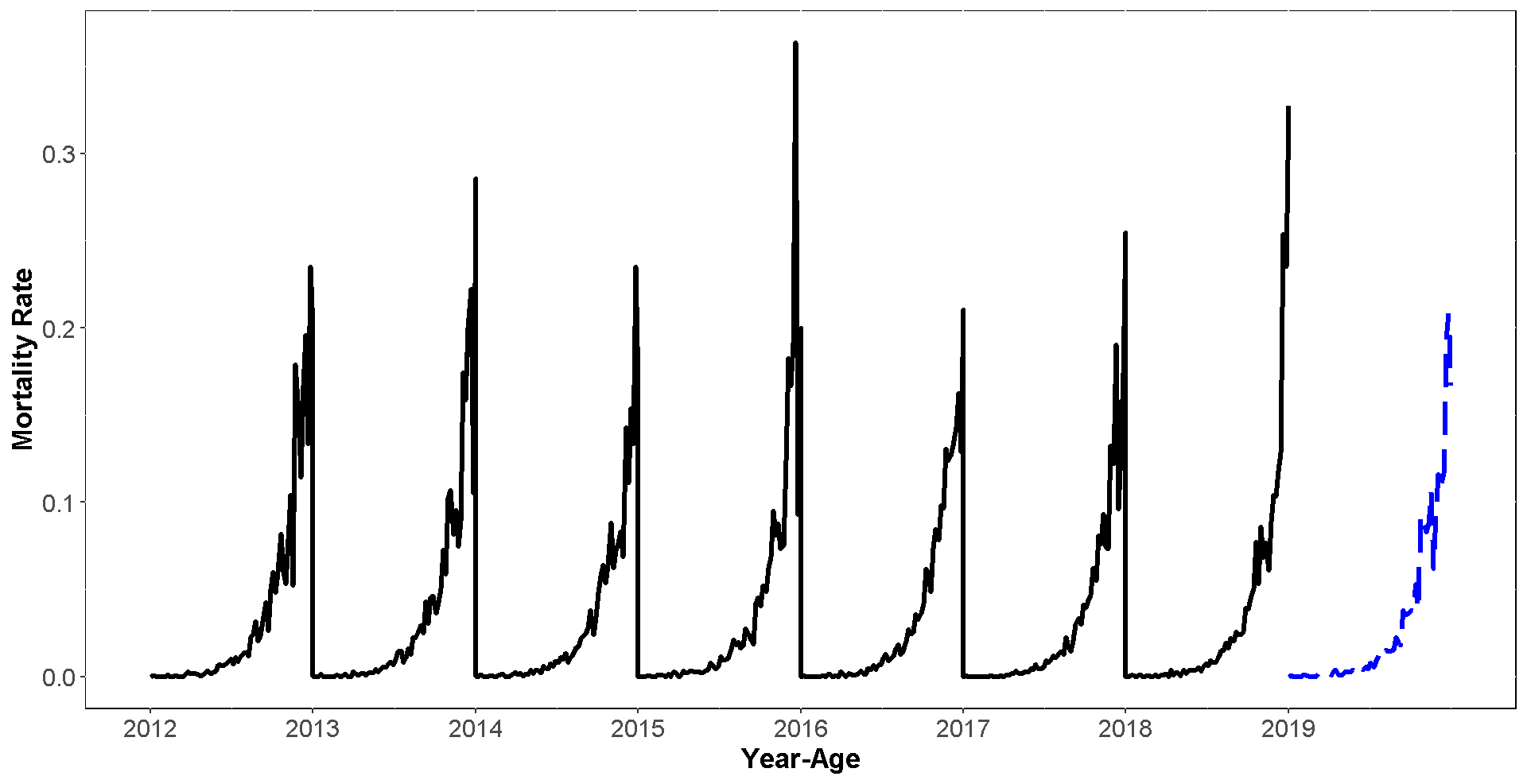}

\label{fig:lstm_serie}
\end{figure}

In LSTM-2 and MHA-2 models, mortality rates are stacked by age and by year, in this order, as illustrated in Figure \ref{fig:lstm_serie}. For the mortality rate at age $x$ in a given year $t$, the mortality rates from previous ages of the same year $t$ and all ages from all previous years to $t$ are taken into account. Thus, the vector of $W_{x,t}^i$ is defined as:

$$W_{x,t}^i = [x, i, t, m_{30:x-1, t}^{i}, m_{30:95, 2012:t-1}^{i}]$$

In the LSTM-3 and MHA-3 models, the mortality rates from the two preceding years ($t-1$ and $t-2$) are included for ages $x-1$ and $x-2$, respectively, for both male and female genders. This approach aims to capture generational effects. The vector \( W_{x,t}^i \) is defined as follows:

\[
W_{x,t}^i = [x, i, t, m_{x-1,t-1}^i, m_{x-2,t-2}^i]
\]

Before using these vectors in training these models, min-max normalization is performed on the covariate and predictor matrices. This step is crucial to ensure that all variables are on the same scale and contribute uniformly to the learning process, preventing large variations in scales between different variables from affecting the model's performance.

The LSTM and MHA networks received the dynamic variables as input. The plots (ii) and (iii) of Figure \ref{fig:arquit} show their basic architecture. The outputs of these two networks were subsequently connected to an FNN network layer to account for the static variables and generate the final output ($m_{x,t}^i$). These architectures can be seen in plot (iv) of Figure \ref{fig:arquit}.

Some adjustments and refinements were made to the neural network models, which are worth mentioning in this section. For the FNN networks, different architectures were tested (number of layers, number of nodes in each layer and activation functions). This paper only shows the results of the best performance architecture. By the way, the FNN showing the best performance had 4 hidden layers (see plot (i) at Figure \ref{fig:arquit}). The ultimate layer had a sigmoid activation function and the loss function was the mean squared error. For the LSTM and MHA networks, the best fit was obtained using a single layer, which received the dynamic variables as input, and in parallel, an FNN network with 4 hidden layers responsible for the static variables. The outputs of these two networks were subsequently connected to an FNN network layer to generate the final output ($m_{x,t}^i$). For both of these networks, the loss function was the mean absolute error.

\subsection{Performance}

To evaluate one-year out-of-sample performance, predicted values were compared with observed values, and the metrics of Mean Absolute Error (MAE) and Root Mean Squared Error (RMSE) were calculated. MAE and RMSE are two metrics that measure the differences between predicted and observed values. They are calculated as follows:

$$MAE = \frac{1}{n} \sum_{i=1}^n | m^i_{x,t^*} - \hat{m}^i_{x,t^*} |$$

$$RMSE = \sqrt{\frac{1}{n} \sum_{i=1}^n (m^i_{x,t^*} - \hat{m}^i_{x,t^*} )^2 }$$

\noindent where $n = \sum_{i,x,t^*} 1$ and $t^*$ is the forecasting year.

\section{Results}

\begin{figure}[!]
\centering
\caption{Schematic representation of the temporal cross-validation procedure. In the first step, the training set, represented by the black nodes, consists of data from  2013 to 2015, and mortality predictions are obtained for 2016, represented by the gray nodes, with performance metrics calculated for this year's data. In subsequent steps, data from the next year is added to the training set, and predictions and metrics are obtained for the subsequent year. This procedure is called time series cross-validation.}
\vspace{0.4cm}

\includegraphics[height=5cm]{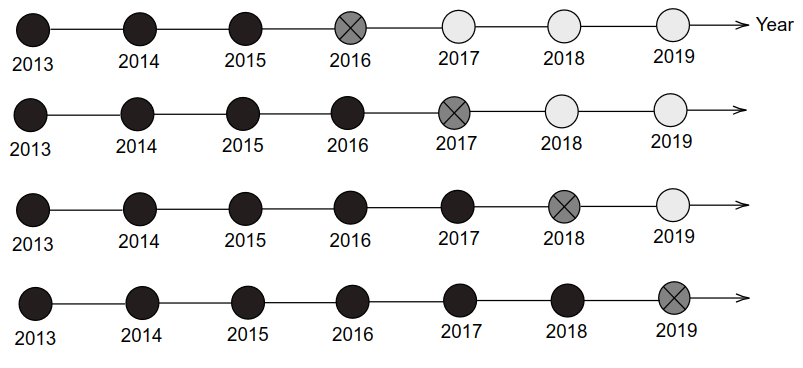}

\label{fig:ts_cv}
\end{figure}

In this section, we present the results of the methods applied to forecast the mortality of our pension fund sample. To avoid misinterpretation of the results, we decided to use data from 2012 to 2019 in a temporal cross-validation approach, with schematics presented in Figure \ref{fig:ts_cv}. The process starts assuming the model is being fit by the end of 2015, which allows the modeler to use data from 2013 to 2015 (inclusive). For this standpoint, the aim is to forecast next year's mortality, so performance metrics (MAE and RMSE) are computed for 2016. This is represented in the first line of Figure 5. For the following year, data from 2016 is also available and the forecasting accuracy is computed for 2017. The performance metrics (MAE and RMSE) for the out-of-sample cross-validation are presented in Table \ref{tab:models}, which shows them for each one of the models testes in this paper. Ultimately, the best performance averages (the lowest ones), based on MAE and RMSE, will represent the best models.

\begin{figure}[!]
\caption{Lee-Carter Model. Observed mortality rates in 2019 (dots) with the respective predicted mortality curve. On the left: Males. On the right: Females.}
\vspace{+0.1in}

\centering

\includegraphics[width=7.40cm]{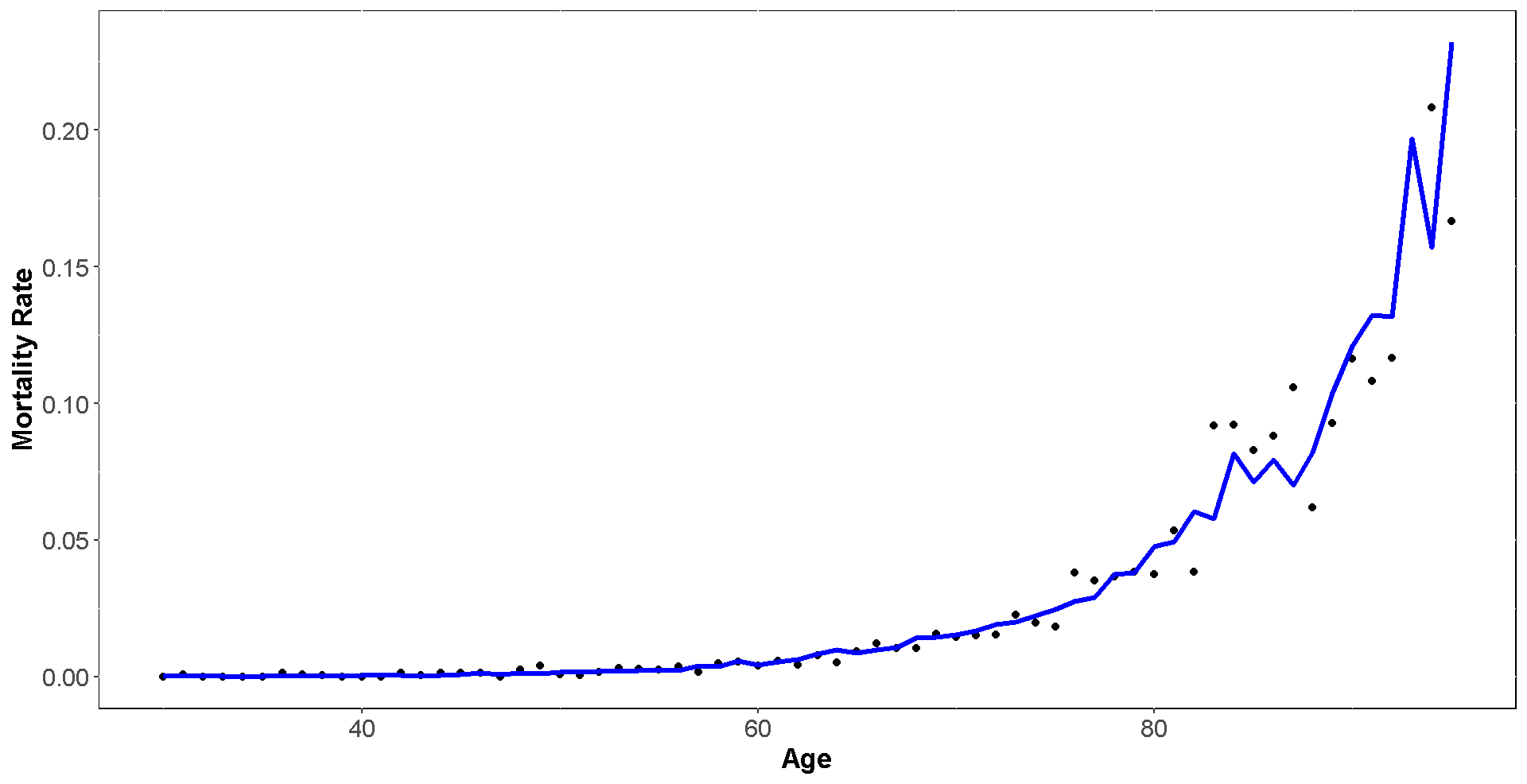}
\includegraphics[width=7.40cm]{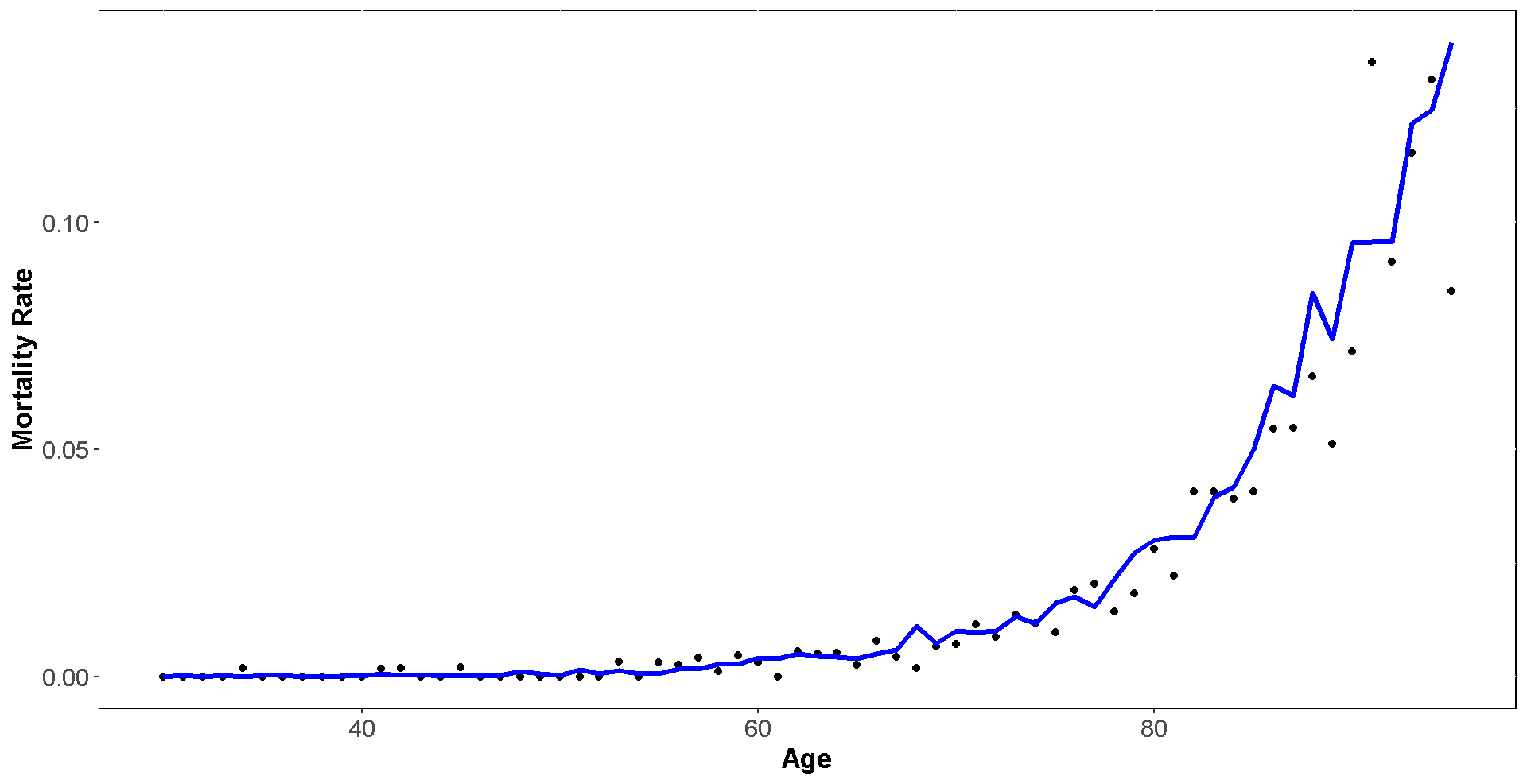}

\label{fig:LC}
\end{figure}

\begin{figure}[!]
\caption{AR - Regression Tree and RF - Random Forest. Observed mortality rates in 2019 (dots) with the respective predicted mortality curve. On the left: Males. On the right: Females. First row: Regression Tree. Second row: Random Forest.}
\vspace{+0.1in}

\centering

\includegraphics[width=7.4cm]{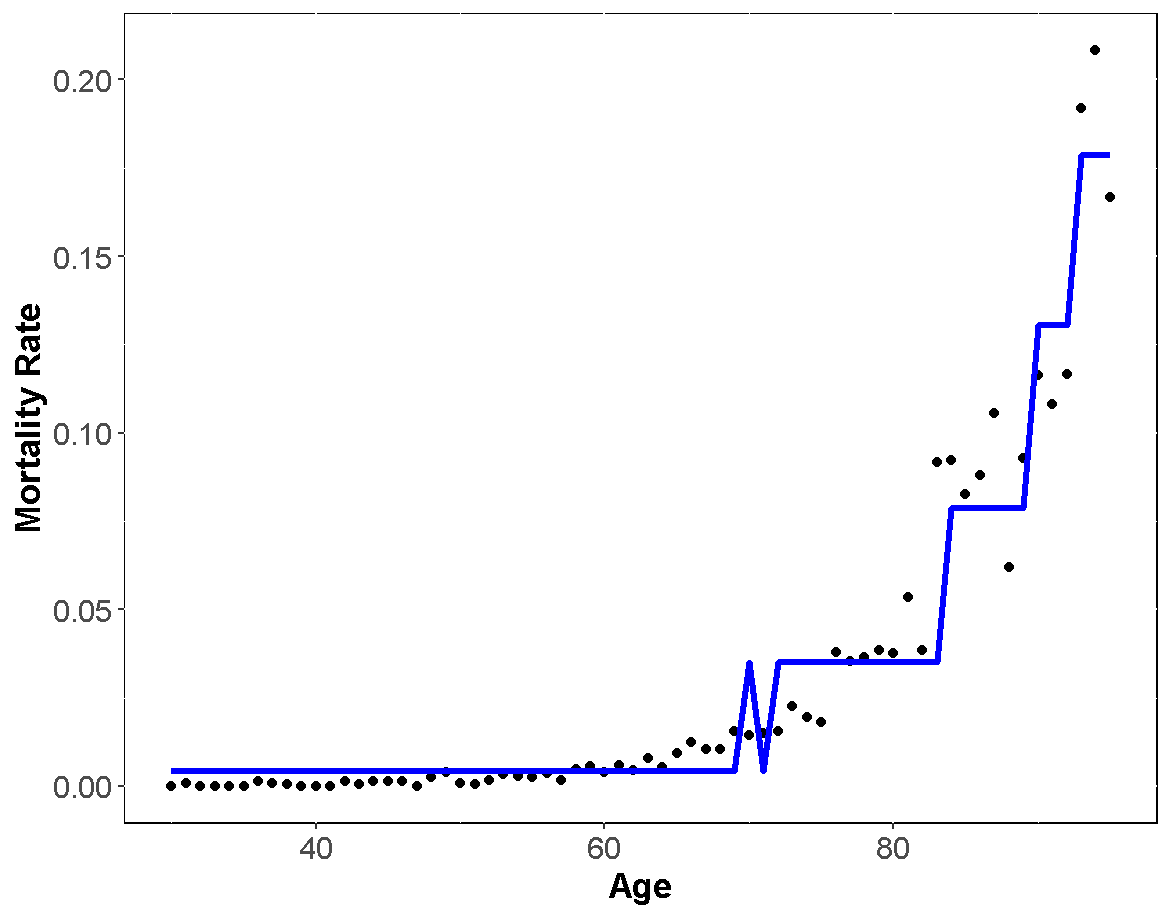}
\includegraphics[width=7.4cm]{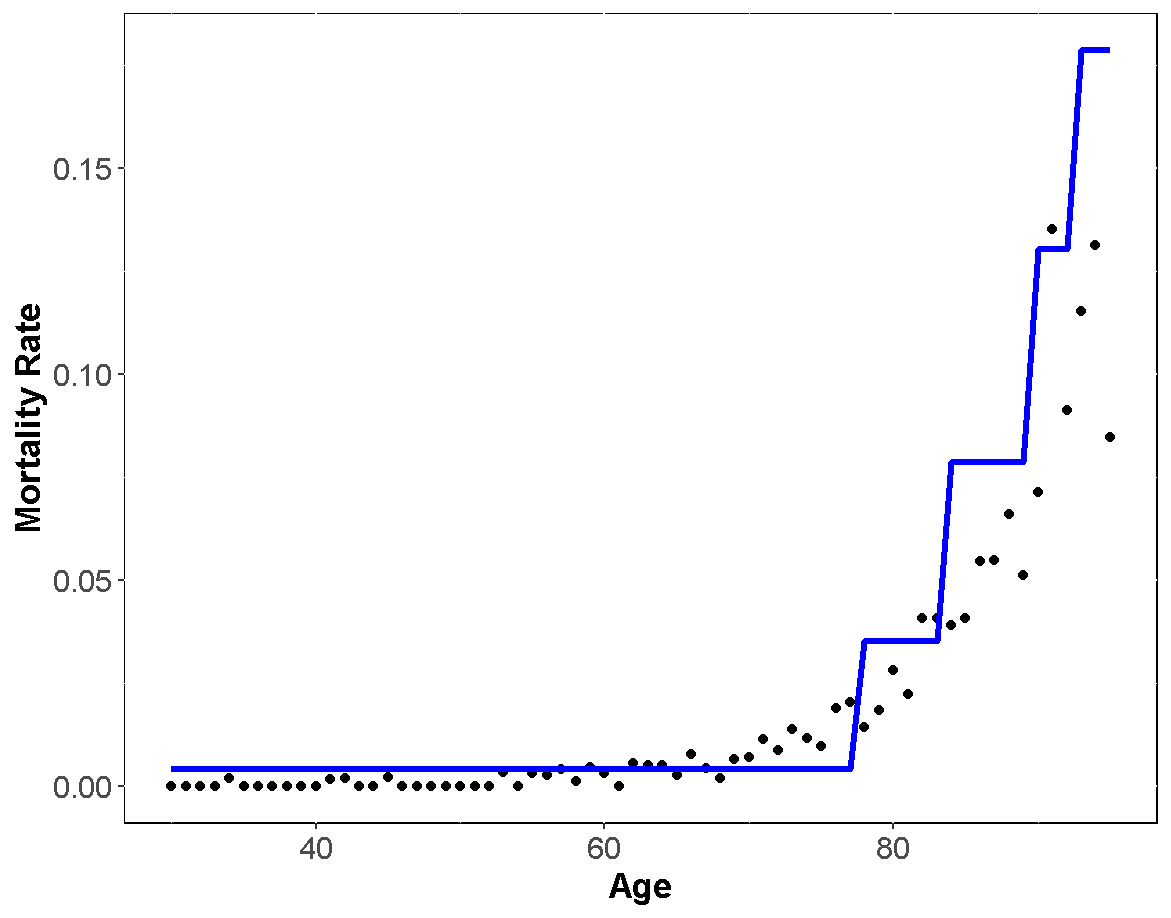}
\includegraphics[width=7.4cm]{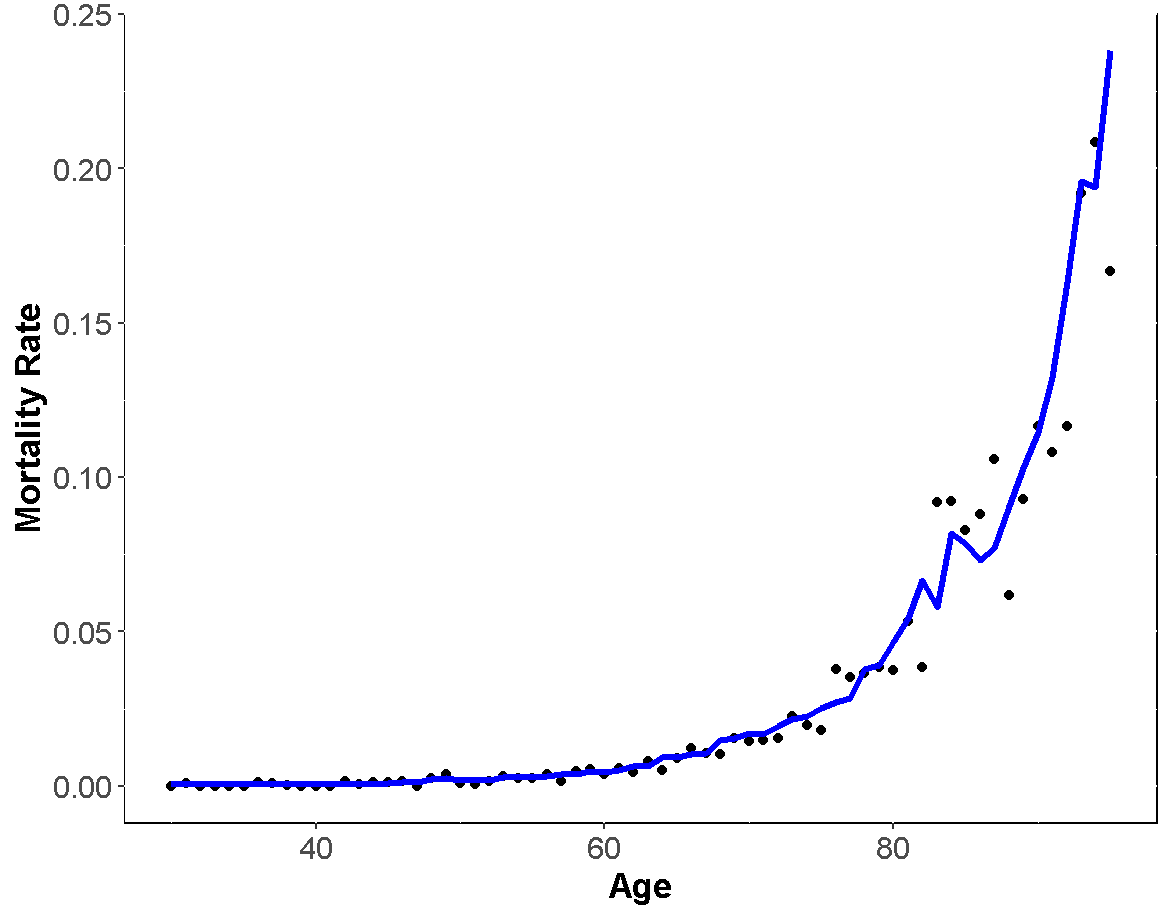}
\includegraphics[width=7.4cm]{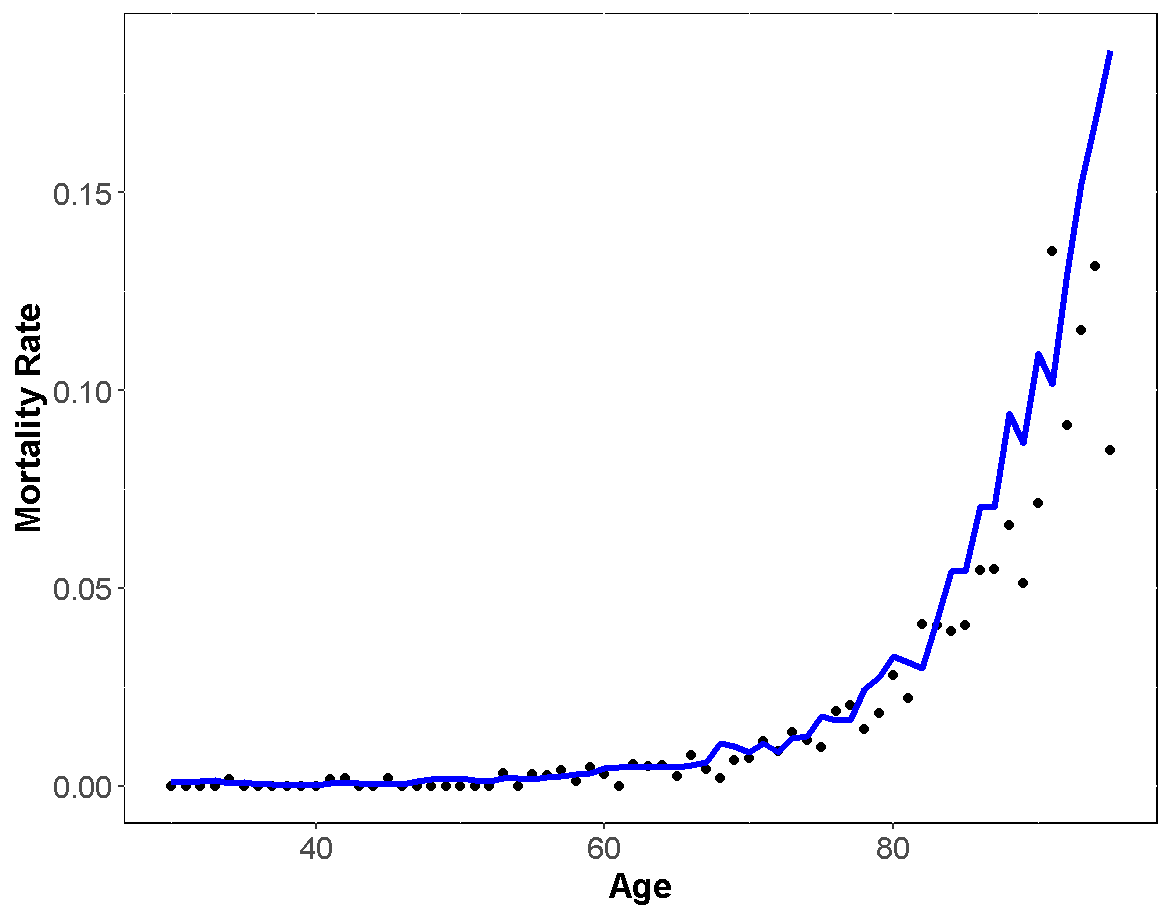}

\label{fig:AR_funds}
\end{figure}

\begin{figure}[!]
\caption{BST - Boosting and XGB - XGBoost. Observed mortality rates in 2019 (dots) with the respective predicted mortality curve. On the left: Males.On the right: Females. First row: Boosting. Second row; XGBoost.}
\vspace{+0.1in}

\centering

\includegraphics[width=7.4cm]{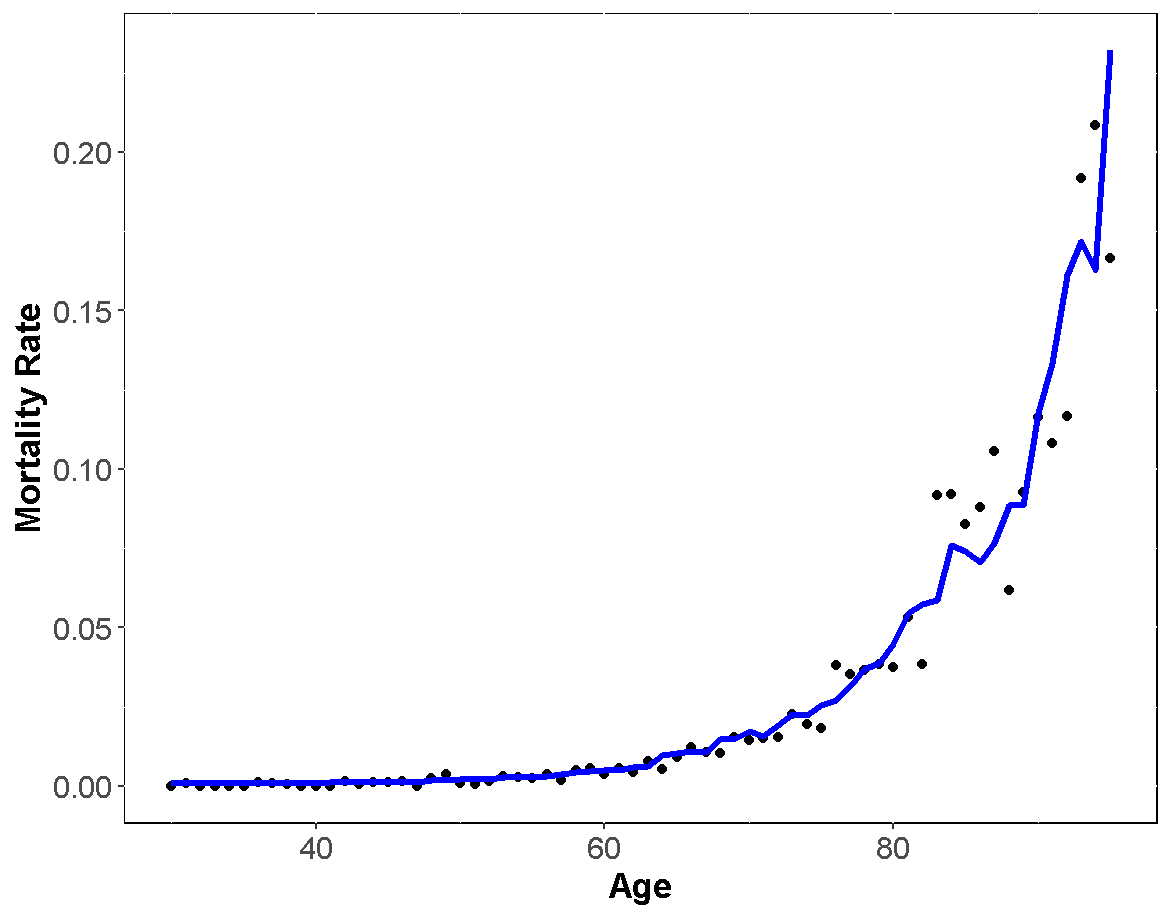}
\includegraphics[width=7.4cm]{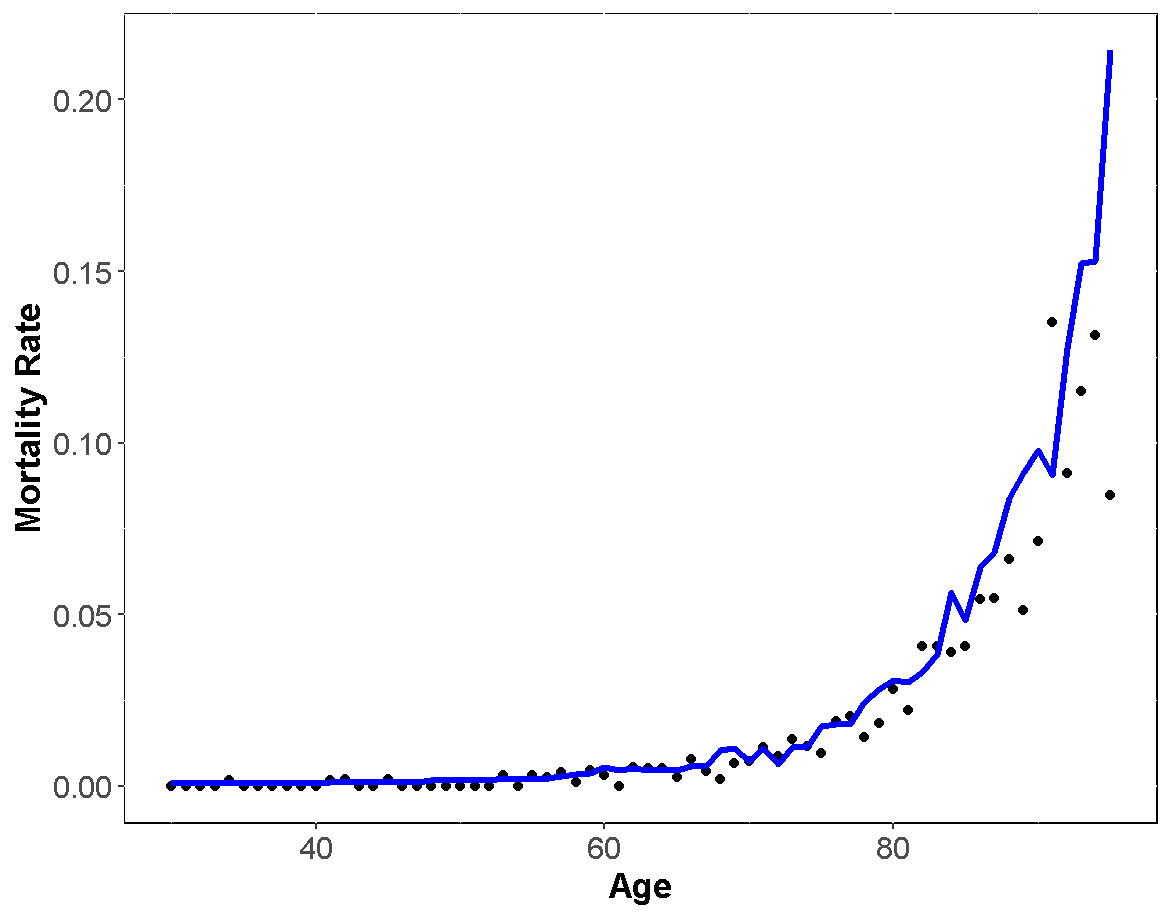}
\includegraphics[width=7.4cm]{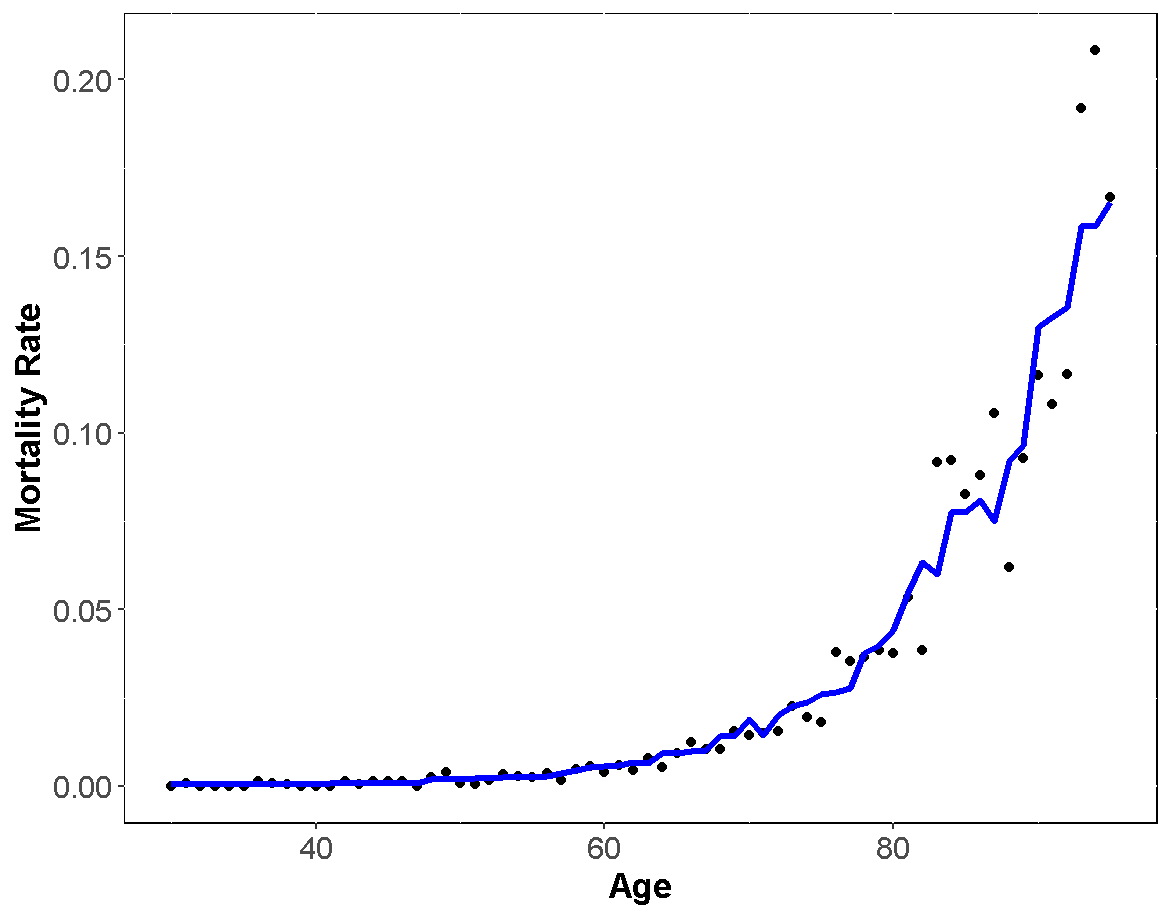}
\includegraphics[width=7.4cm]{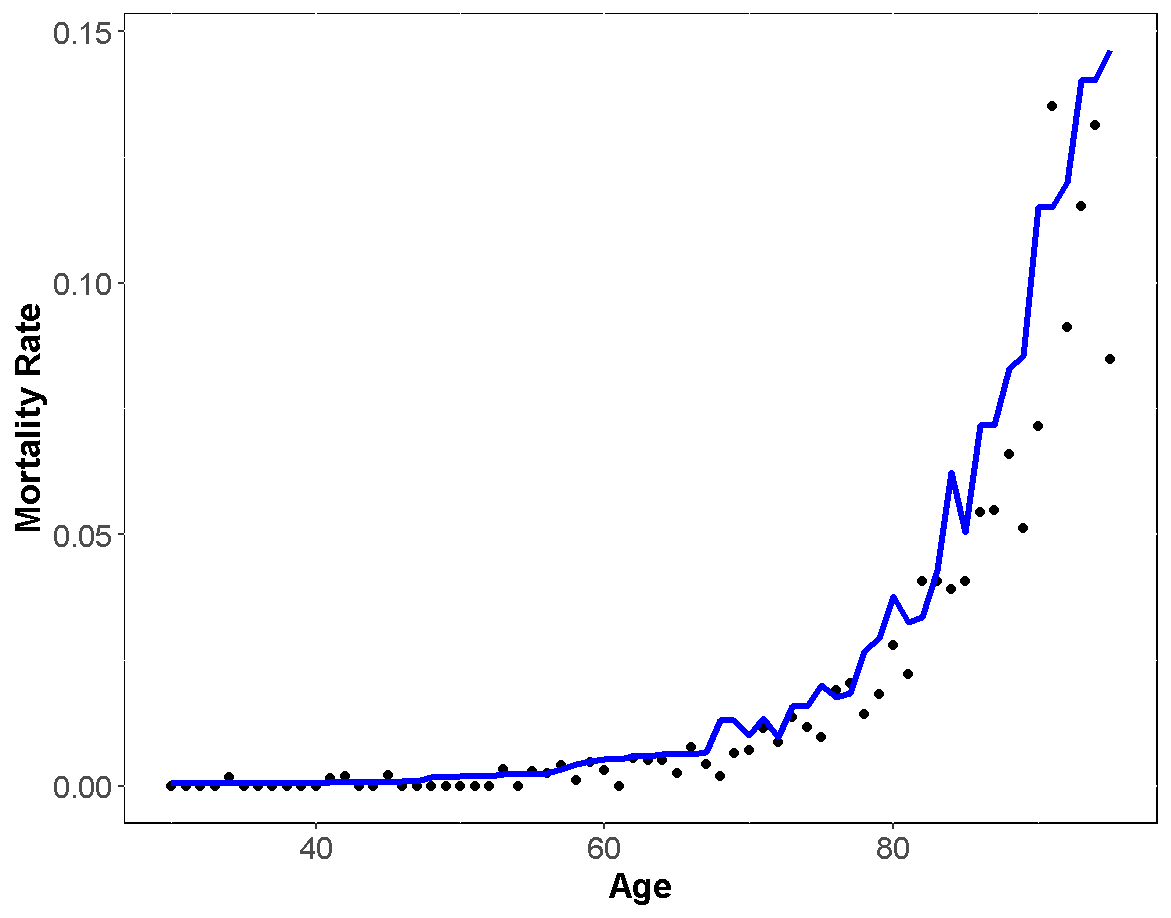}

\label{fig:XGB_funds}
\end{figure}

\begin{figure}[!]
\caption{FNN - Feedforward Neural Network. Observed mortality rates in 2019 (dots) with the respective predicted mortality curve. First row: Males. Second row; Females.}
\vspace{+0.1in}

\centering

\includegraphics[width=7.4cm]{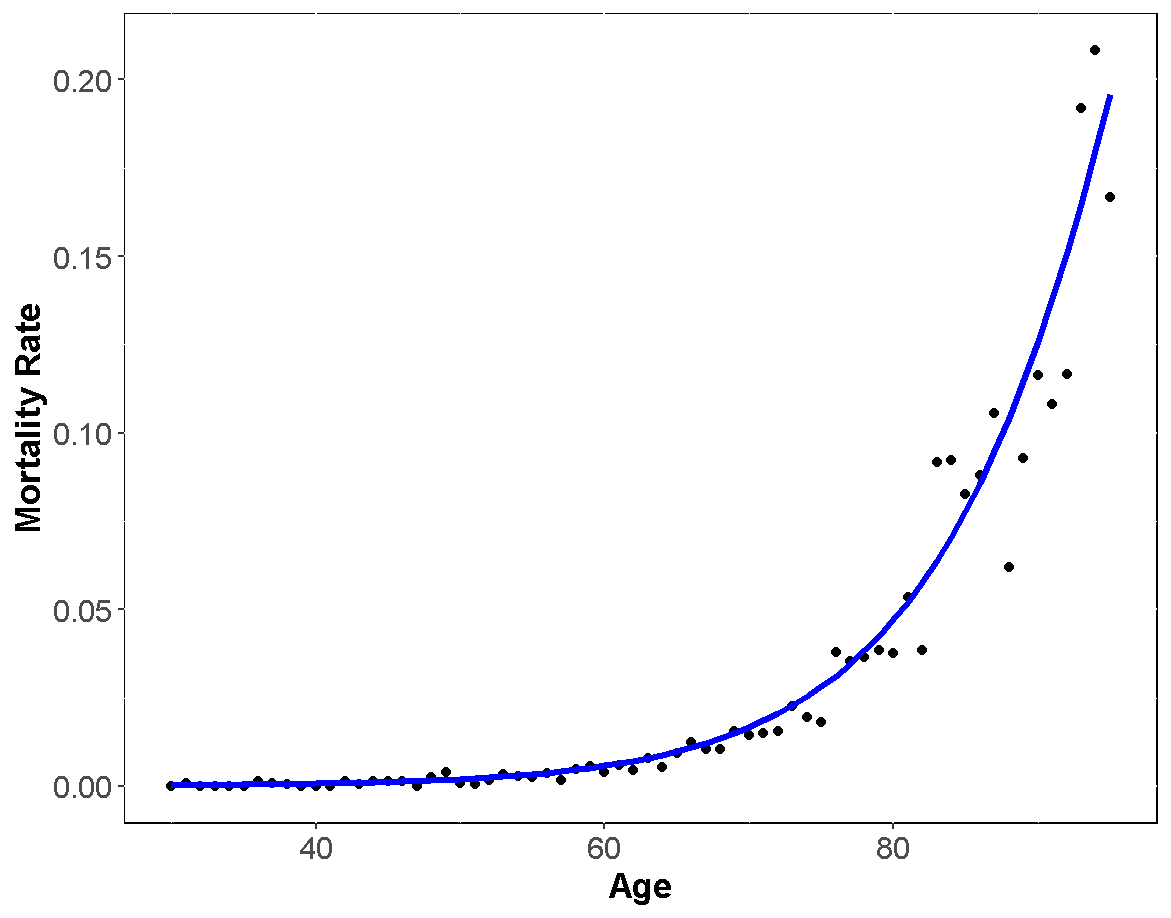}
\includegraphics[width=7.4cm]{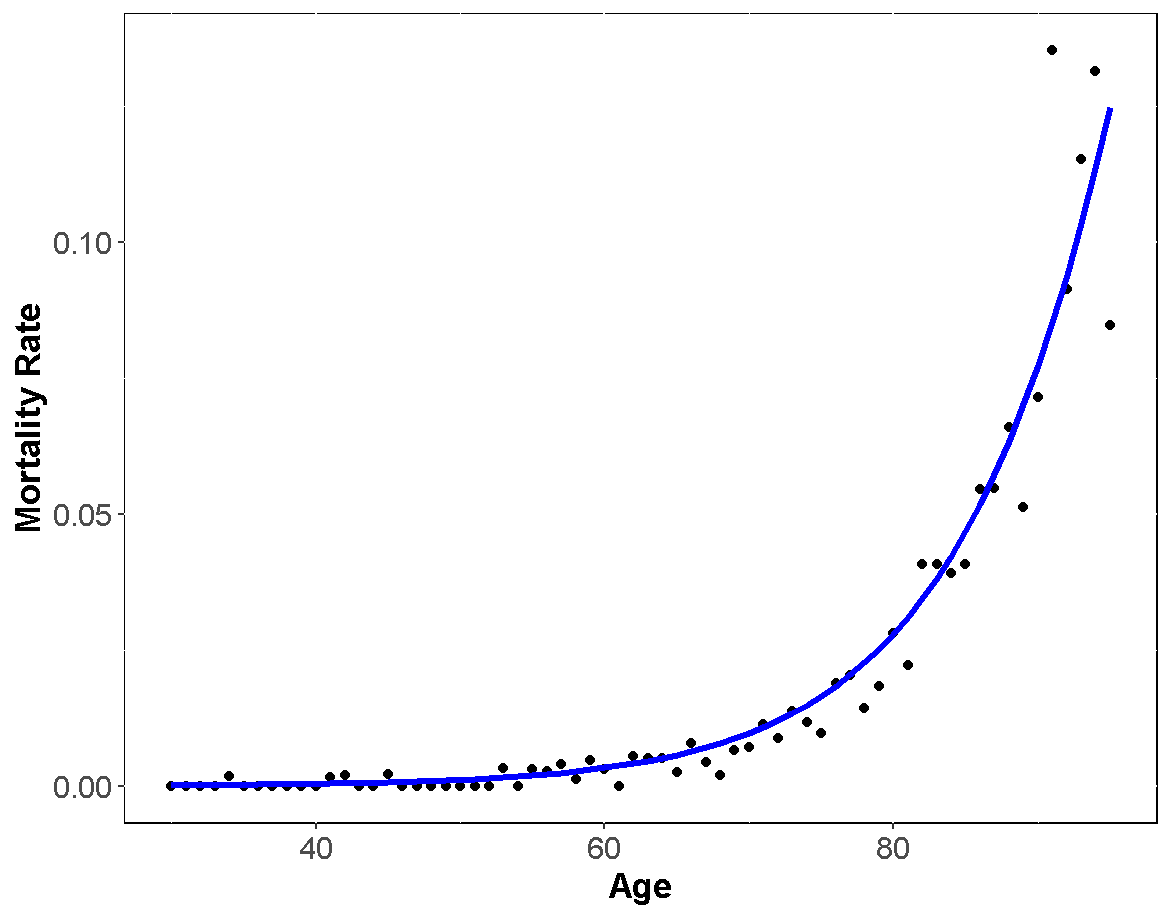}

\label{fig:FNN_funds}
\end{figure}

\begin{figure}[!]
\caption{CatBoost, LSTM, and MHA. Observed mortality rates in 2019 (dots) with the respective predicted mortality curve for the Male gender. Order: (i) CatBoost (ii) LSTM-1 (iii) MHA-1 (iv) LSTM-2 and (v) MHA-2.}
\vspace{+0.1in}

\centering
\begin{tabular}{cc}
\includegraphics[width=7.4cm]{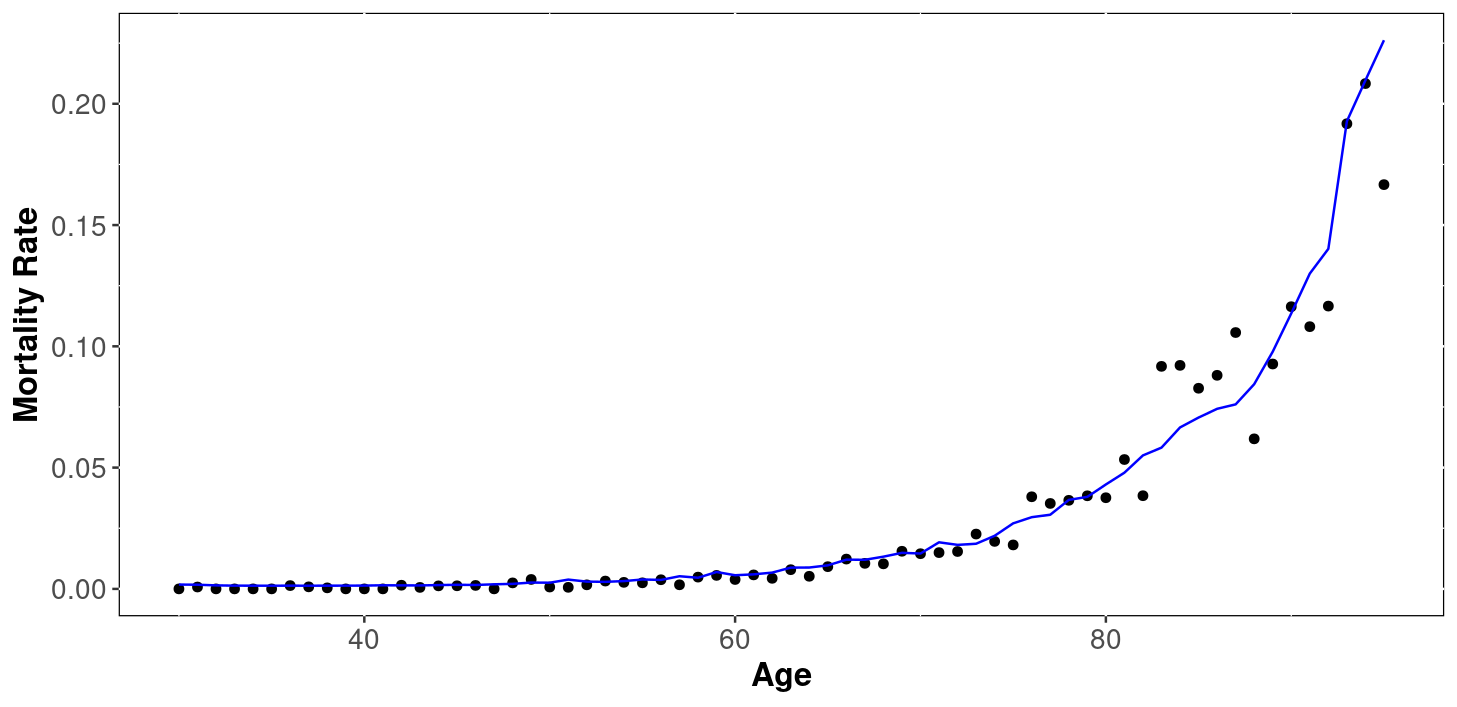} & \includegraphics[width=7.4cm]{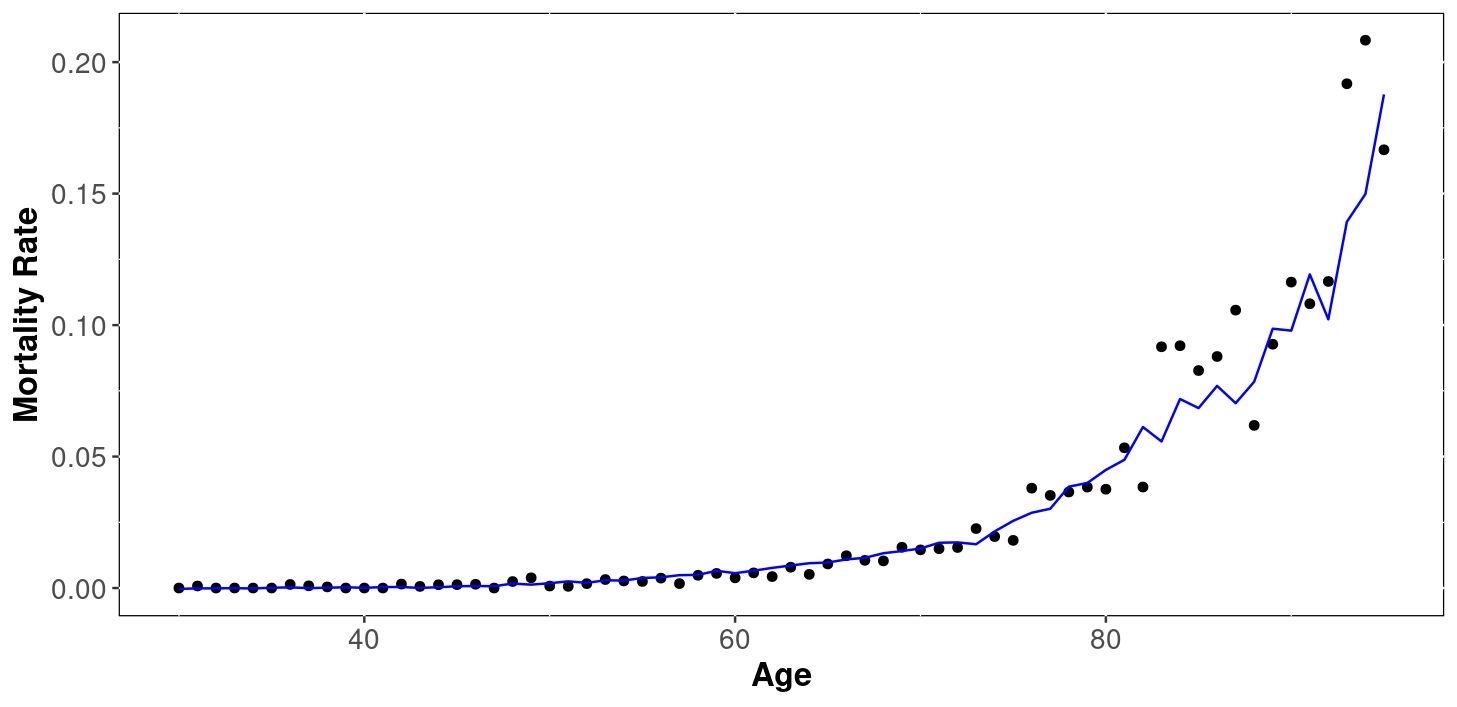}\\
CatBoost    & LSTM-1 \\ 
\includegraphics[width=7.4cm]{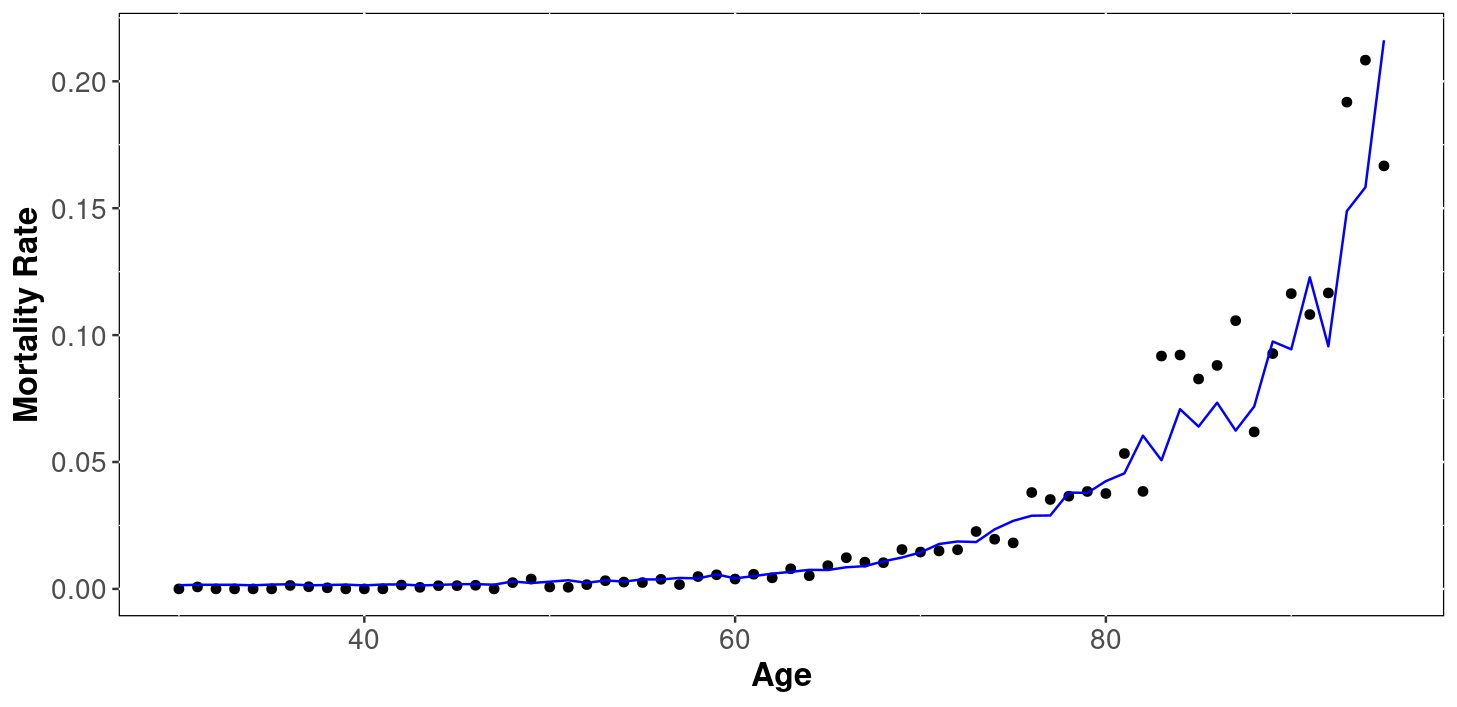} &
\includegraphics[width=7.4cm]{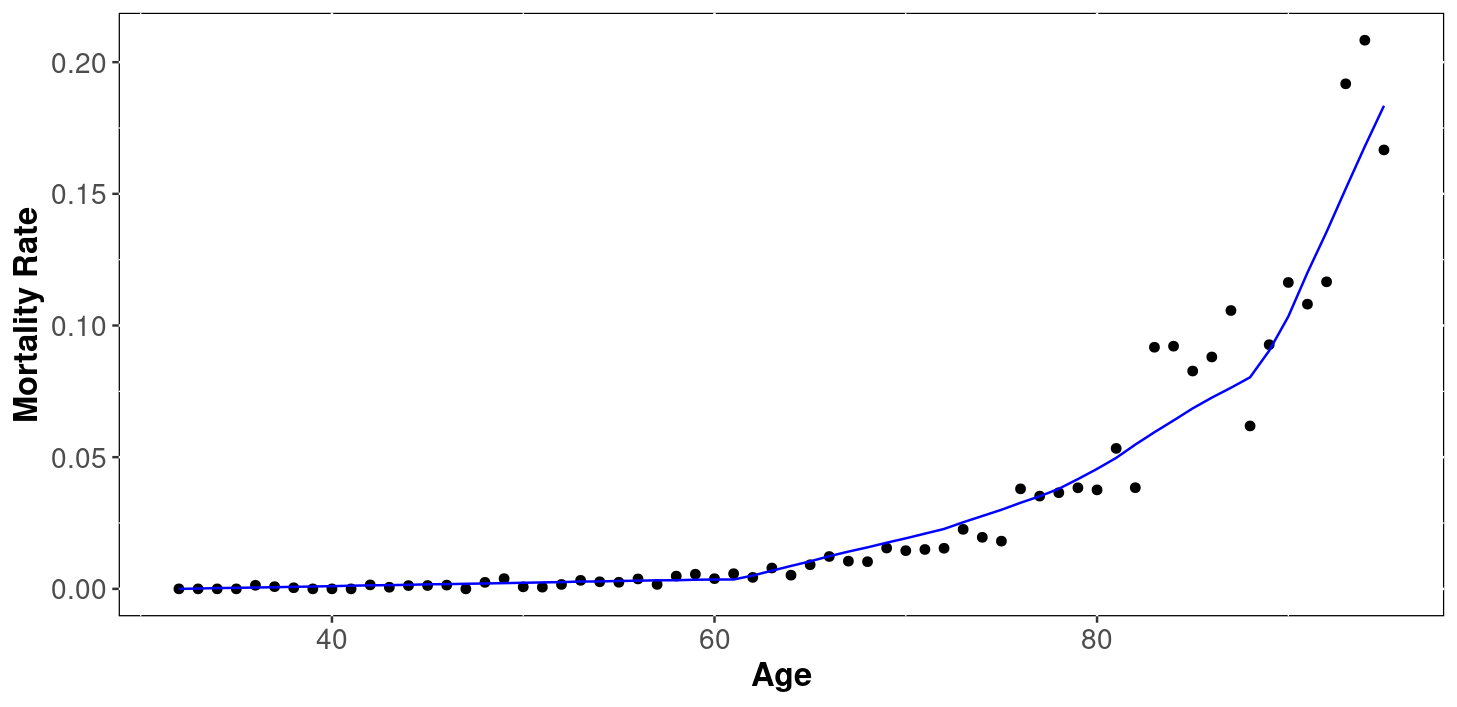} \\
MHA-1 & LSTM-2 \\
\includegraphics[width=7.4cm]{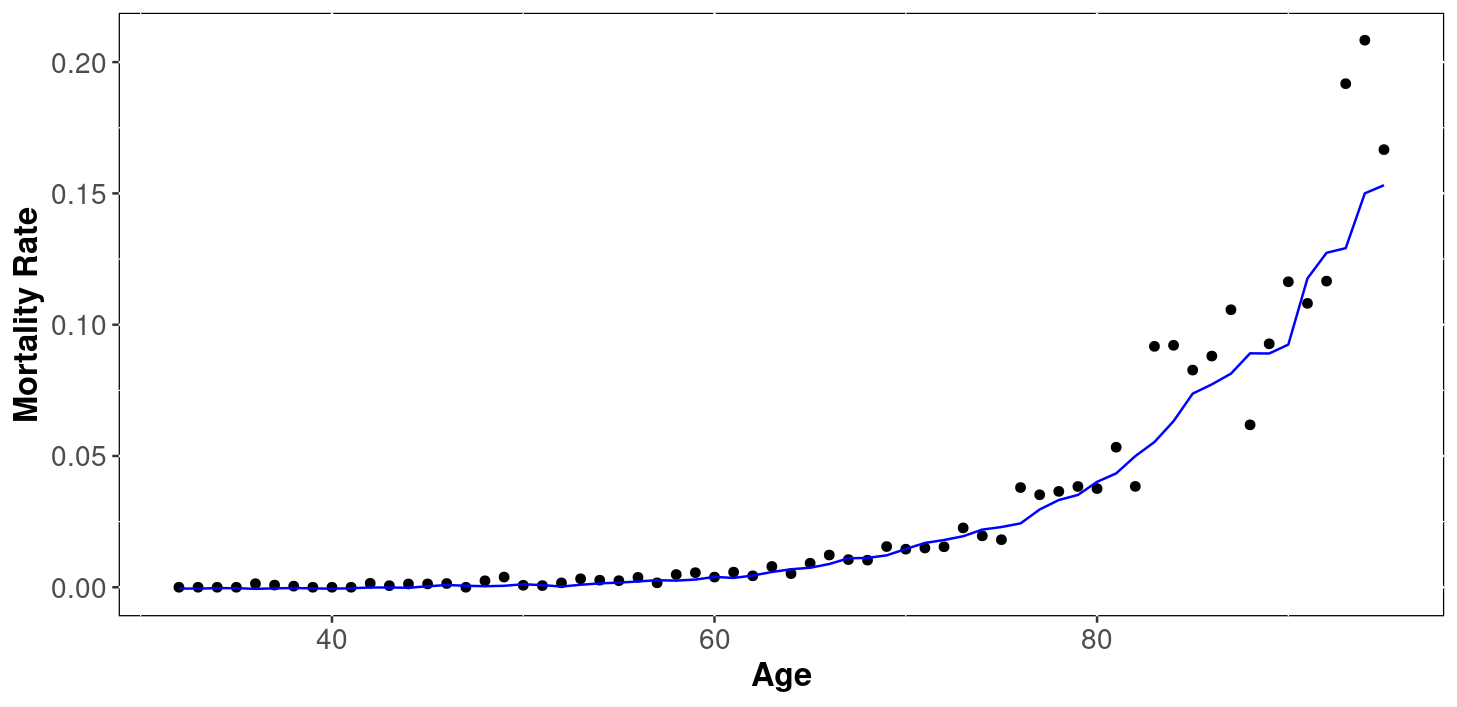} & \\
MHA-2 & \\
\end{tabular}

\label{fig:cat_lstm_mha}
\end{figure}

Figures \ref{fig:LC} -- \ref{fig:cat_lstm_mha} present realized mortality rates for 2019 (black dots) and the mortality curves (in blue) predicted by each one of the models when using data up to 2018.

\begin{table}
\centering
\begin{tabular}{|c|c|c|}
\hline

Model & MAE & RMSE \\ \hline

RT         & 0.0104 & 0.0190 \\
RF         & 0.0068 & 0.0159 \\
BST        & 0.0071 & 0.0182 \\
XGB        & 0.0066 & 0.0159 \\
Catboost   & {\bf 0.0065} & 0.0148 \\
FNN        & 0.0069 & {\bf 0.0138} \\
LSTM-1     & 0.0069 & 0.0162 \\
MHA-1      & 0.0069 & 0.0165 \\
LSTM-2     & 0.0066 & 0.0164 \\
MHA-2      & 0.0066 & 0.0164 \\
LSTM-3     & 0.0066 & 0.0165  \\
MHA-3      & 0.0068 & 0.0170  \\
Lee-Carter & {\bf 0.0065} & 0.0157 \\ \hline

\end{tabular}
\caption{\label{tab:models} Out-of-sample results considering the applied methods. Sample: 2012-2019. Mean of performance metrics for cross-validation. Ages: 30-95.}
\end{table}

As shown in Table \ref{tab:models}, CatBoost and Lee-Carter achieved roughly the same out-of-sample MAE, which is smaller than any other model tested. When the performance metric is chosen as the RMSE, the best model was the FNN, followed closely by Catboost. It should also be mentioned that even though none of the models impose any structure on the resulting mortality curves, the FNN results in a highly desirable smooth and strictly increasing curve. Additionally, it is observed that models based on LSTM and MHA networks (where the mortality rate series were loaded in a ``stacked'' manner), and XGBoost were quite competitive when compared to the benchmark Lee-Carter model.

Figure \ref{fig:LC} shows the mortality projection plots for 2019 and the predicted mortality curve based on the traditional Lee-Carter model. Figures \ref{fig:AR_funds} and \ref{fig:XGB_funds} present the plots of machine learning models (regression tree, random forest, boosting, and XGBoost). It is possible to observe the characteristics of the predicted mortality curves when tree-based algorithms are used: curves produced are not smooth.

Figure \ref{fig:FNN_funds} shows the plots of the FNN. For this methodology, out-of-sample results with a smoothing characteristic are obtained. Based on the plot, an apparent good fit for raw mortality rates may also be seen for both males and females.

Despite the performance metrics being similar when comparing Lee-Carter to ML algorithms, one should note that FNN produced a 2019 out-of-sample predicted mortality curve smoother than the others. This is an important feature when modeling mortality even for sub (and selected) populations that are inherently smaller than national populations. These results made us choose this algorithm to perform applications in the next section.

\begin{figure}[!]
\caption{Out-of-sample absolute residuals heat maps for FNN and Lee-Carter models. Years: 2016-2019. First row: FNN. Second row: Lee-Carter.}
\vspace{+0.1in}
\centering
\includegraphics[width=16.5cm]{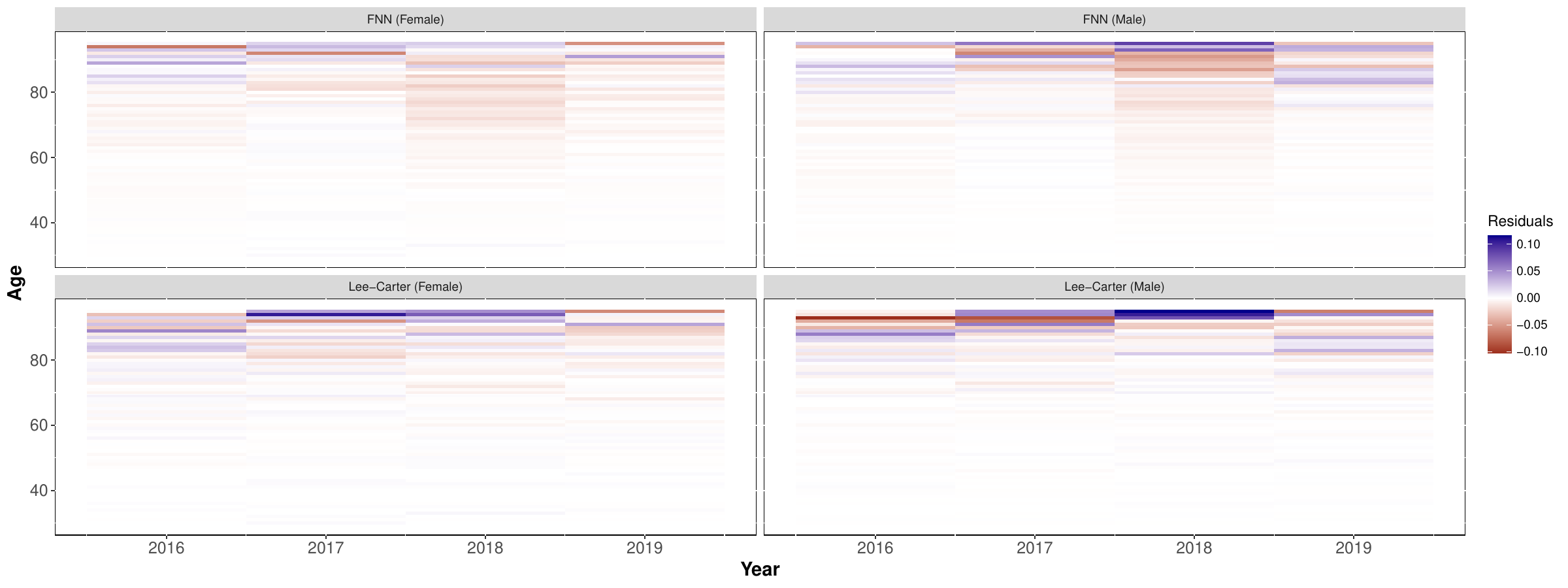}

\label{fig:heatmap}
\end{figure}

Moreover, regarding the residuals heat maps presented in Figure \ref{fig:heatmap}, they show that fitting in younger ages is better than in older ones ($80+$) for both models (FNN and Lee-Carter) and both genders. It is an expected result since older ages have much less data (exposure). For the same reason, one may also notice that the fit for Males presents lower residuals for both models when compared to Females. The Lee-Carter fit presents higher absolute residuals than FNN for Males. For Females, the FNN fit presents more symmetric residuals than the Lee-Carter one. In summary, the plots in Figure \ref{fig:heatmap} also show a better performance of FNN for one-year ahead forecasting in the sample period.

\section{Applications}

\begin{figure}[!]
\caption{Exposure for years 2020 and 2021 - Males: observed (black line) x predicted (red line) by FNN neural network.}
\vspace{+0.1in}
\centering
\includegraphics[width=12cm]{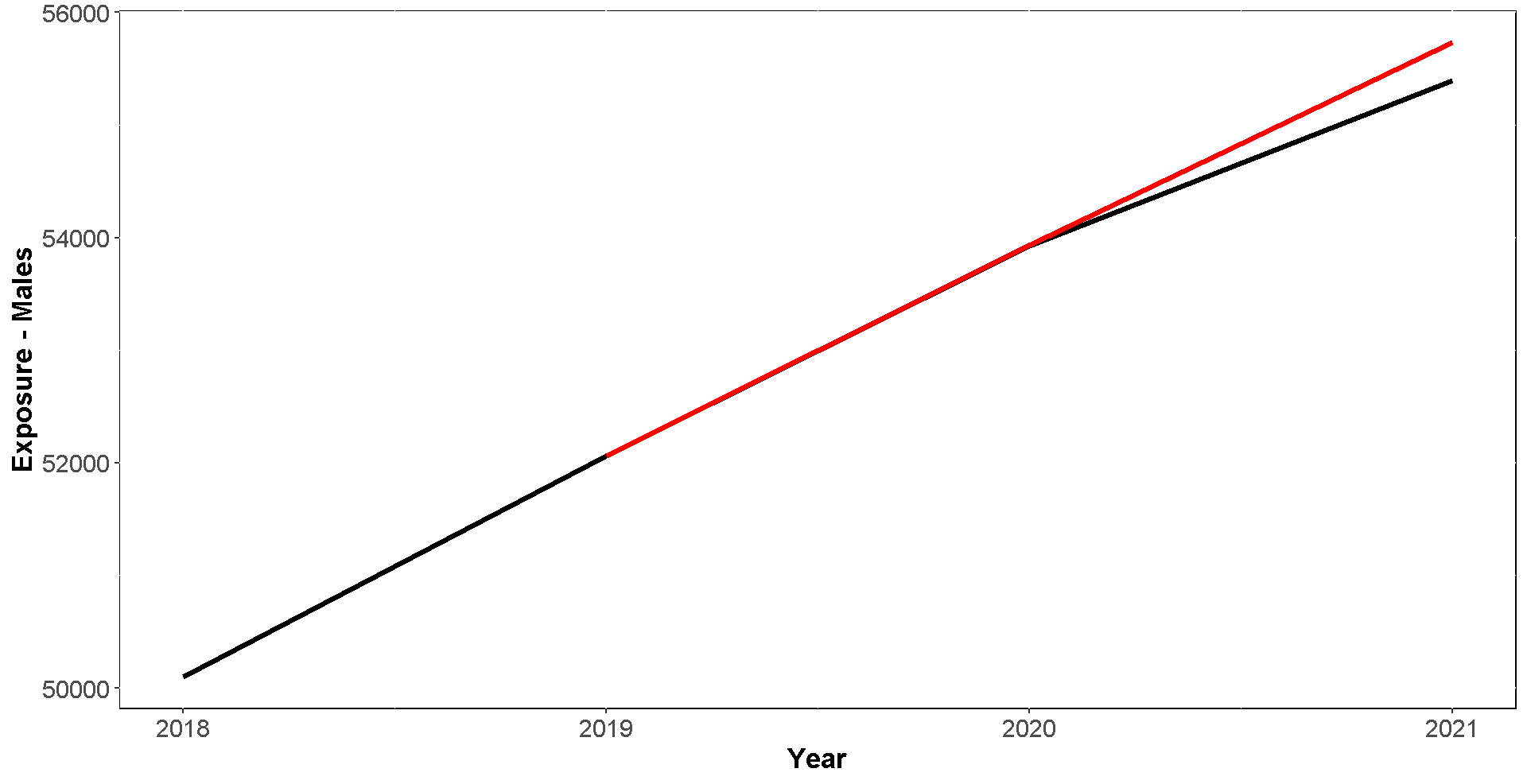}
\label{fig:pandemics}
\end{figure}

\begin{figure}[!]
\caption{Projected cash flow for the years 2022 to 2031 (10 years) of the pension fund sample population at ages 60 to 95 in 2021, considering: (i) the FNN method, and (ii) the 2021 insurance market mortality table (BR-EMS 2021).}
\vspace{+0.1in}
\centering
\includegraphics[width=12cm]{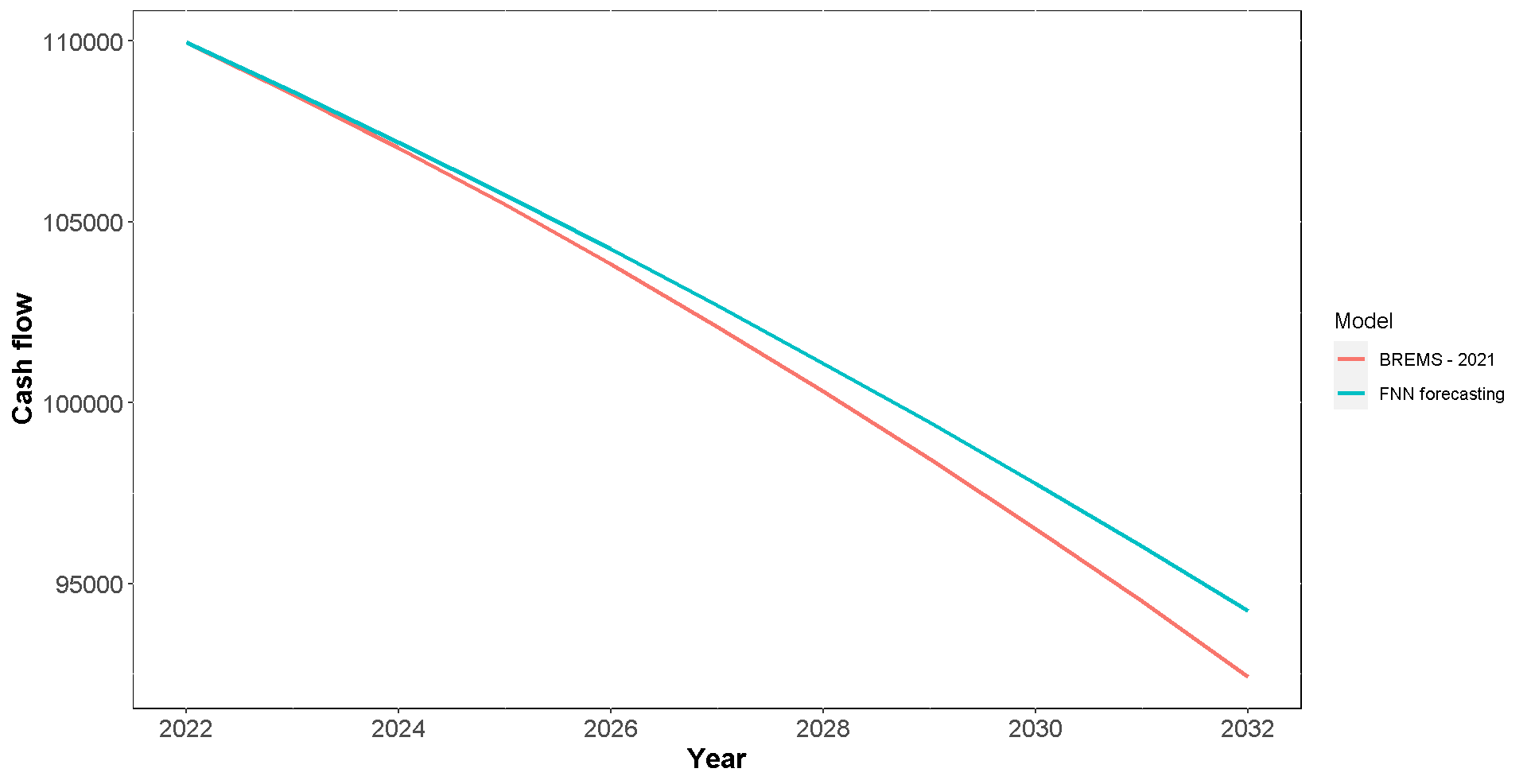}
\label{fig:exp_pred}
\end{figure}

For practical application purposes, we will consider in this section the FNN, which obtained the best RMSE in Table \ref{tab:models}. As applications of the results, we: (i) forecasted life expectancy at age 60 over time - Table \ref{tab:e60}, (ii) estimated the effect of the pandemics on the pension fund sample in the years 2020 and 2021 - Figure \ref{fig:pandemics}, (iii) measured a hypothetical mathematical provision to retirees over 60 years old in the sample, considering \$1 of annual income, and (iv) constructed the expected cash flow for income granted to pensioners over 60 years old in 2021 for the following 10 years - Figure \ref{fig:exp_pred}. Such cash flow is a necessary input for asset and liability management (ALM) or market risk calculation (specifically, mismatch risk) purposes. 

\begin{table}
\centering
\begin{tabular}{ c|c|c  }
Year & $e_{60}$ Male & $e_{60}$ Female \\ \hline
2022 & 24.89 & 27.86 \\
2023 & 25.06 & 28.01 \\
2024 & 25.24 & 28.17 \\
2025 & 25.42 & 28.32 \\
2026 & 25.60 & 28.47 \\ \hline
\end{tabular}
\caption{\label{tab:e60} FNN Neural Network. Forecast of life expectancy at age 60 ($e_{60}$). Training sample: 2012-2019. Forecast period: 2022-2026. Ages: 60-100.}
\end{table}

The mathematical provision calculated consistently with current and realistic mortality assumptions, considering \$1 of annual income and 5\% per year of real interest, was \$ 376,825. For comparison purposes, if the BR-EMS 2021 mortality table\footnote{Available in http://www.susep.gov.br/setores-susep/cgpro/copep/Tabuas\%20BR-EMS\%202010\%202015\%202021-010721.xlsx} for Males' survival is used, the value is \$ 370,210 (a difference of 1.76\%).

Based on the projection made for the years 2020 and 2021, with data up to 2019, it is observed that there is a predicted exposure value higher than the observed one for ages over 60 years for the Male gender. The difference was 0.5\%. The five-year age group that showed the highest relative difference was from 80 to 84 years.

Finally, exposures of the study population for the next 10 years (2022 to 2031) for each gender were projected. Based on the year 2021, exposures for ages 60 to 95 were projected for the years 2022 to 2031 using expected future mortality rates obtained with the FNN neural network model. These projected exposures can provide future cash flows for current retirees. Figure \ref{fig:exp_pred} illustrates the total annual future exposures considering the best fit (FNN) and also the fixed mortality table from the insurance industry for 2021 (BR-EMS 2021). The projected exposures are consistent with the results presented in this article. According to the trained models, the mortality of the pension fund population is lower than that predicted in the insurance industry table for 2021.

\section{Concluding remarks}

A consistent evaluation of pension fund obligations requires the use of consistent, realistic, and updated mortality rates and also the prediction of these mortality rates over time for cash flow discounting purposes. Most mortality rate projection studies use national population data when performing applications. Despite the basis risk, the use of national mortality rates may not pose a significant problem when the difference between the national population and the selected population is not relevant. However, in countries with large social inequalities, mortality rates can be quite different between a selected population and the average national population. In Brazil, national mortality rates are higher than those of a selected subpopulation, such as pension fund participants or insurance customers.

In this paper, several machine learning methods and neural networks were applied to predict the mortality of participants over 30 years old from a pension fund population. The use of machine learning in actuarial science has been termed actuarial learning. The methods used in this paper were decision tree, random forest, boosting, XGBoost, CatBoost, FNN, LSTM and MHA neural networks. We compared the results obtained with the Lee and Carter (1992) model, a widely used benchmark for mortality rate forecasting purposes. Our results show that actuarial learning models are a competitive alternative for mortality rate forecasting for a selected population. Using RMSE as the performance metric, the best fit was achieved using the FNN neural network. If MAE is considered, there are some competitive models with Lee-Carter, such as CatBoost, LSTM, and MHA.

The applications made in the article indicate, among other things, the differentiation in mortality levels between Males and Females, the longevity improvement in the evolution of life expectancy over the years, and an estimated number of deaths for 60+ in the years 2020 and 2021 that may be due to the COVID pandemic. Furthermore, we also highlight the forecasting of cash flows consistently, which is a fundamental tool for pension fund risk management, as it is a necessary step for ALM risk evaluation (called ``mismatch risk'').

\vspace{+1.0in}

\noindent The code supporting the findings of this study is available on request from the corresponding author, EFLM. The data are not publicly available since they belong to some Brazilian pension funds that were only made available to the authors through a non-disclosure agreement - NDA.

\bibliographystyle{apalike}
\bibliography{references}

\ \ 

\end{document}